\pdfoutput=1

\documentclass[11pt]{article}

\usepackage{ACL2023}

\usepackage{times}
\usepackage{latexsym}
\usepackage{booktabs}
\usepackage[T1]{fontenc}
\usepackage{hyperref}
\usepackage[utf8]{inputenc}
\usepackage{graphicx}
\usepackage{amsmath}
\usepackage{microtype}

\usepackage{inconsolata}
\usepackage{multirow}

\newcommand{\name}{~ConceptPsy}
\newcommand{\NEEM}{National Post-graduate Entrance Examination}

\title{\name: A Benchmark Suite with Conceptual Comprehensiveness in Psychology}

\author{Junlei Zhang$^{1,2*}$\quad 
Hongliang He$^{1,2*}$\quad 
Nirui Song$^{2}$ \quad
Zhanchao Zhou$^{2}$ \quad
Shuyuan He$^{2}$ \quad \\
\textbf{Shuai Zhang}$^{1,2}$ \quad
\textbf{Huachuan Qiu}$^{1,2}$ \quad
\textbf{Anqi Li}$^{1,2}$ \quad
\textbf{Yong Dai}$^{\ddagger}$ \quad
 \textbf{Lizhi Ma}$^{2\dagger}$ \quad
\textbf{Zhenzhong Lan}$^{2\dagger}$ \quad \\
   \quad $^1$Zhejiang University \quad $^2$School of Engineering, Westlake University \\
  \texttt{\{zhangjunlei,malizhi,lanzhenzhong\}@westlake.edu.cn}\\
  }

\begin{document}
\maketitle

\renewcommand{\thefootnote}{\fnsymbol{footnote}}
\footnotetext[1]{Equal Contribution.}
\footnotetext[2]{Corresponding author.}
\footnotetext[3]{Independent Researcher.}
\renewcommand{\thefootnote}

\begin{abstract}

The effective incorporation of Large Language Models (LLMs) into the field of psychology necessitates a comprehensive domain-specific benchmark to guide their development and adaptation. Existing Chinese benchmarks in the style of MMLU, such as C-EVAL and CMMLU, do include psychology subjects, but their concept coverage is far from exhaustive. The number of questions in each domain is just in the hundreds, and an uneven question sampling process can lead to a ``concept bias'' issue. This bias, stemming from using a question set with a low concept coverage rate to represent a subject, can potentially lead to skewed results. To address this, we present \name, a unique benchmark specifically designed for evaluating Chinese LLMs' complex reasoning and knowledge in psychology. \name~ encompasses 12 core subjects and 1,383 concepts from official exams. To avoid copyright issues, we prompt \texttt{GPT-4} to generate questions for each of the concepts, which are then validated by psychology professionals to ensure high quality. Besides the overall scores, we annotate each question with a chapter label to provide fine-grained results. We evaluate a range of LLMs on \name~ and the results show significant performance differences across psychology concepts, even among models from the same series. We anticipate the comprehensive concept coverage and the fine-grained strengths and weaknesses identified by \name~ can facilitate the development and growth of the Chinese psychology domain.
\end{abstract}
\section{Introduction}
\begin{figure}[t!] 
\centering 
\includegraphics[width=0.5\textwidth]{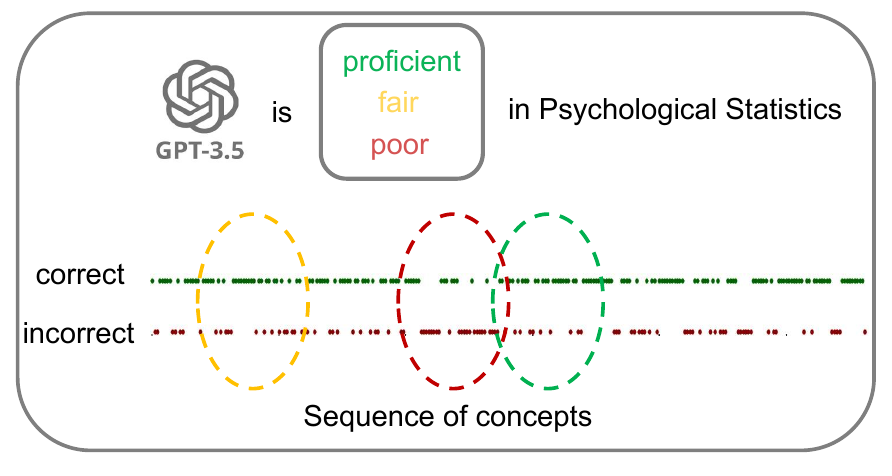} 
\caption{\texttt{GPT-3.5-Turbo}'s concept-wise performance on Psychological Statistics. The x-axis represents the sequence of concepts, arranged in the order they appear in the textbook. The dashed circles represent sampled question sets. Different samplings can mislead people's understanding of a model.}
\label{fig: score of different concepts} 
\end{figure}

\begin{figure*}[t!] 
\centering 
\includegraphics[width=\textwidth]{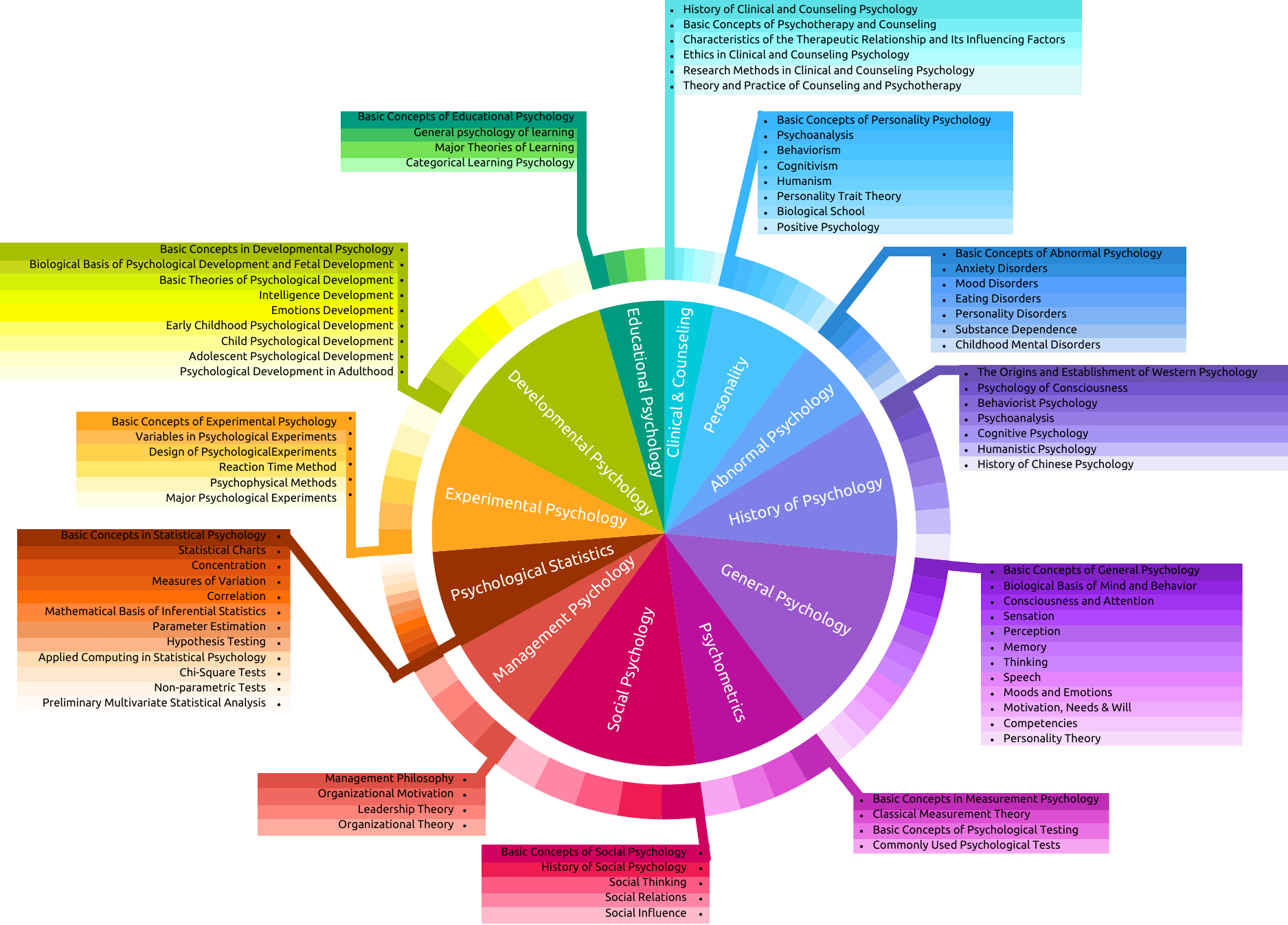} 
\caption{Diagram overview of concepts in \name. We sample questions based on the requirement of the National Post-graduate Entrance Examination in China. Each question is tagged with a modified chapter name, serving as the chapter-level concept, to further provide chapter-level accuracy. }
\label{fig: intro_of_conceptpsy} 
\end{figure*}

Domain-specific benchmarks are essential for the advancement and fine-tuning of LLMs.  Recently, there has been a significant advancement in LLMs, exemplified by models like \texttt{GPT-4} ~\citep{achiam2023gpt}, \texttt{Claude-3}~\citep{claude3}, \texttt{Qwen2}~\citep{qwen}, \texttt{DeepSeek-V2} ~\citep{deepseekv2}. These models have shown remarkable abilities in a wide array of standard tasks. However, a critical gap remains in the systematic understanding of these models' performance within the important domain of psychology. This gap primarily arises from the absence of a comprehensive psychological benchmark.

While some Chinese MMLU-style benchmarks such as CMMLU~\citep{li2023cmmlu}  have offered a wide range of subjects for assessment. They tend to focus more on covering many subjects rather than diving deep into each one. This is reflected by the limited number of questions per subject and the lack of emphasis on concept coverage during the sampling process. As a result, these questions do not thoroughly cover the necessary concepts within each subject.  We term this problem as ``concept bias'' and discuss more in \S \ref{sec: Analysis on the Concept Coverage}. Relying on question sets with limited conceptual breadth to represent a subject can lead to misleading results. Figure \ref{fig: score of different concepts} illustrates the performance variability of models across concepts, with each question set distinguished by a unique color. These variations highlight that models can exhibit a range of proficiency—from poor to fair to proficient—based on the specific questions used for evaluation.  As a result, it might mislead developers about how well a model can handle psychology-related topics.

Domain benchmarks are key to advancing LLMs and have been introduced in fields like \citep{fei2023lawbench}, Finance \citep{zhang2023fineval}, Mathematics \citep{wu2024conceptmath}, Medicine \citep{chen2024benchmarking}. However, a comprehensive Chinese benchmark for evaluating advanced knowledge and reasoning abilities in psychology is missing.  To narrow the gap, we introduce \name, the first comprehensive Chinese evaluation suite designed to thoroughly assess LLMs' psychology knowledge and reasoning abilities. To ensure the comprehensiveness of concepts, we manually gathered 1383 college-level concepts based on the official requirements of \NEEM.  To avoid copyright issues, we design prompts inspired by the Chinese Academy of Sciences Psychological qualification exam. We prompt GPT-4 to generate four multiple-choice questions per concept, which are reviewed by professional psychologists. We also assigned a chapter-level label to each question, providing fine-grained performance analysis to facilitate model improvements.

We conducted experiments evaluating a wide array of LLMs. Results show GPT-4 surpassing human performance with an average accuracy of 84\%. Chinese models like \texttt{Yi-34B-Chat} and \texttt{Qwen1.5-72B-Chat} also scored highly, though disparities in performance across chapters reached up to 50\%. Even within the same series, large-scale models displayed significant chapter-specific performance variations. We believe \name~ will help developers better comprehend their models' psychology abilities and foster the development of foundational psychology models for Chinese users.

\begin{figure}[!t] 
\centering 
\includegraphics[width=0.5\textwidth]{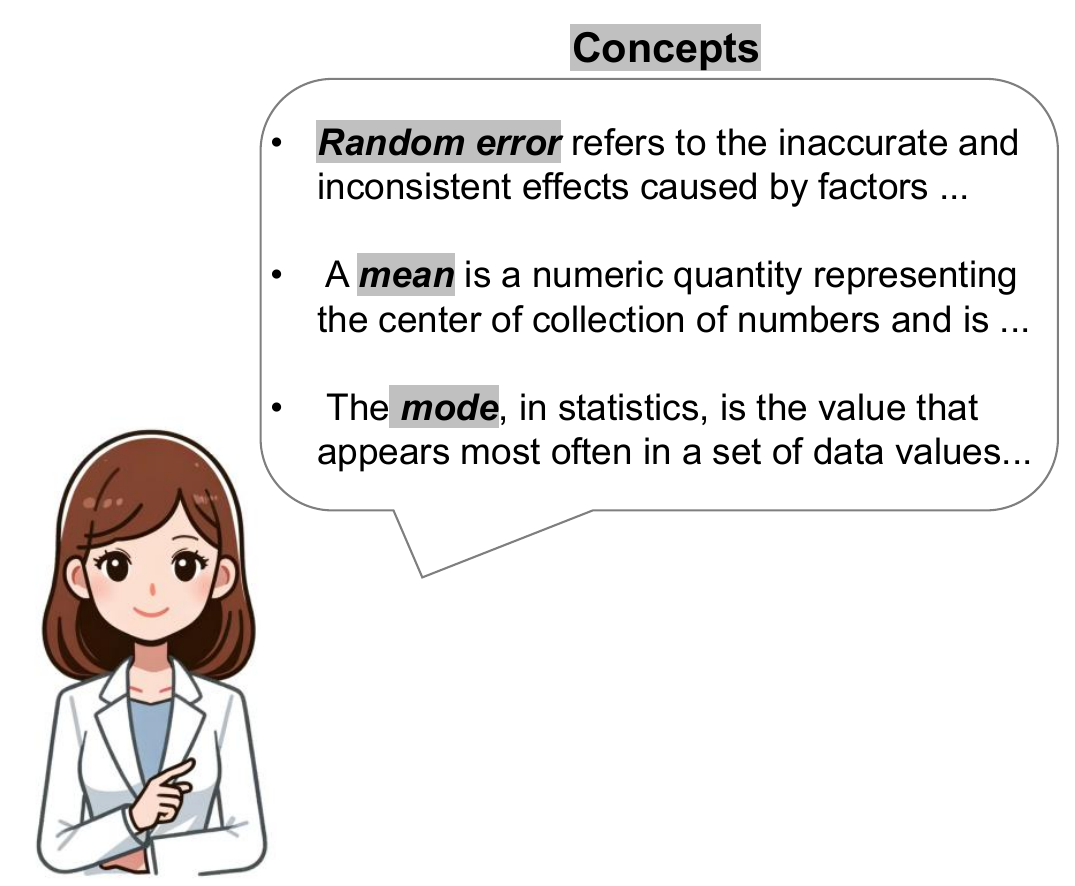} 
\caption{Examples of concepts. We define a ``concept'' as fundamental units of understanding that encapsulate specific knowledge within a broader field of study.}
\label{fig: illustration_of_concept} 
\end{figure}

\begin{figure*}[t!] 
\centering 
\includegraphics[width=\textwidth]{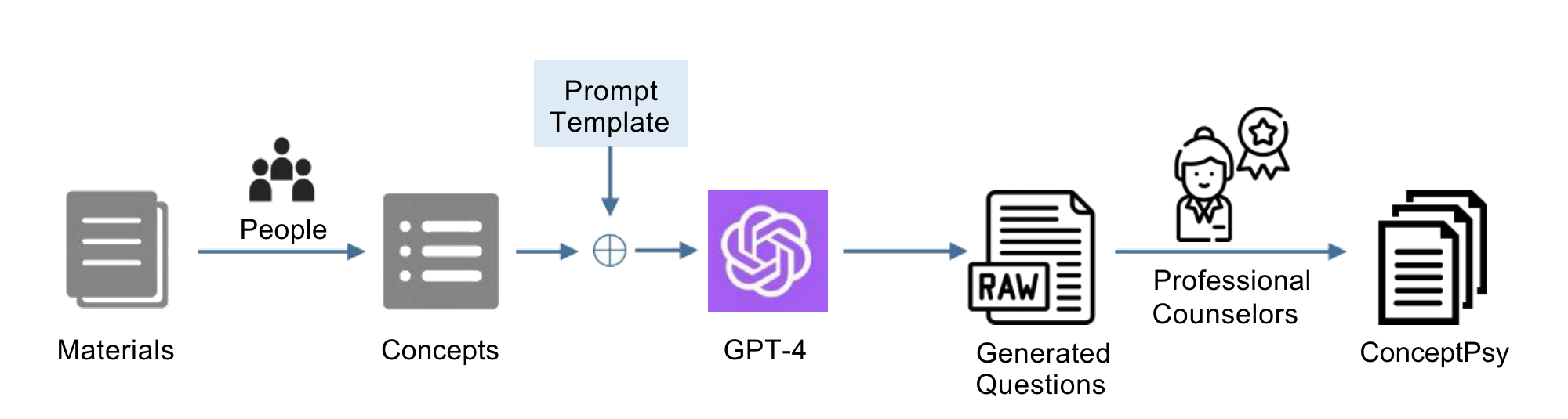} 
\caption{Overview of Our Concept-Driven Framework. We collect relevant concepts based on the requirements of corresponding examinations. To diversify the types of questions, we summarize three question patterns from these exams and design specific prompts for each type. Questions are then generated using \texttt{GPT-4}. Subsequently, we hire professional psychological counselors to review the questions for accuracy and relevance. }

\label{fig: generation pipeline} 
\end{figure*}

\begin{table}[t!]
\centering
\resizebox{0.3\textwidth}{!}{\begin{tabular}{@{}lc@{}}
\toprule
Subject & Coverage Rate \\ \midrule
\multicolumn{2}{c}{STEM} \\ \midrule
Computer Architecture & 0.55 \\
Computer Network & 0.31 \\
High School Biology & 0.52 \\ \midrule
\multicolumn{2}{c}{Social Science} \\ \midrule
High School Geography & 0.48 \\
Marxism & 0.70 \\ \midrule
\multicolumn{2}{c}{Humanity} \\ \midrule
High School History & 0.76 \\
Logic & 0.47 \\ \midrule
Avg & 0.54 \\ \bottomrule
\end{tabular}}
\caption{The concept coverage rate of randomly sampled subjects, with chapter-level concepts generated by \texttt{GPT-4}, as evaluated on C-EVAL.}
\label{tab:concept coverage rate of other subjects}
\end{table}


\begin{table}[t!]
\centering
\resizebox{0.5\textwidth}{!}{\begin{tabular}{@{}lcccc@{}}
\toprule
Subject & Benchmark & \#Questions & \#Concepts & Coverage\_rate \\ \midrule
\multirow{2}{*}{Professional Psychology} & C-EVAL & - & - & - \\
                   & CMMLU & 232 & 84 (chapter-level) & 0.59 \\ \midrule
\multirow{2}{*}{Advanced Math} & C-EVAL & 173 & 94 & 0.54\\
                   & CMMLU & 104 & 36 & 0.35 \\ \bottomrule
\end{tabular}}
\caption{The concept coverage rate of subjects with manually collected required concepts. We prompt \texttt{GPT-4} to classify each question into one or more concepts and subsequently calculate the coverage rate.}
\label{tab: converage rate}
\end{table}


\section{Case Study of Concept Bias\label{sec: Case Study on Chinese MMLU benchmarks}}
We conduct a case study to analyze the concept bias in popular Chinese MMLU benchmarks C-EVAL~\citep{huang2023ceval} and CMMLU~\citep{li2023cmmlu}. We define a concept as a fundamental unit in the human learning process. For example,  in Figure \ref{fig: illustration_of_concept}, the  ``random error''  is a concept when studying ``random variables''. We study concept bias from two perspectives: 1. concept coverage rate (\S \ref{sec: Analysis on the Concept Coverage}): the proportion of concepts tested in the sampled questions to the total concepts required. 2. performance variance (\S \ref{sec: concepts_variance}): the difference in model performance across various concepts.

\subsection{Analysis on the Concept Coverage \label{sec: Analysis on the Concept Coverage}}

In our research, we introduce the term ``concept coverage rate'' to quantify the extent to which the tested concepts in a given question set align with the required concepts in a given subject. Specifically, we calculate it as the ratio of the number of tested concepts, denoted as $C_{\text{collection}}$, to the number of required concepts, denoted as $C_{\text{requirement}}$:  $(Concept_{coverage\_rate} = \frac{C_{collection}}{C_{requirement}}$. C-EVAL excludes subjects related to psychology. Within CMMLU, a subject termed ``professional psychology" contains 232 questions, and the average score is utilized to gauge a model's proficiency in psychology. Our analysis delves into the concept coverage rate specific to psychology within CMMLU. Additionally, we conduct sampling across diverse disciplines to investigate the potential presence of concept bias in other subjects.



\paragraph{Setup} We first assess the concept coverage of ``professional psychology'' in C-MMLU. Utilizing \NEEM, we manually gather all necessary college-level concepts (1383 in total). We then employ \texttt{GPT-4} to assign each question to one or more concepts. To investigate concept bias in other subjects, we sample subjects from various disciplines (STEM, Social Science, Humanities) in C-EVAL. Due to the resource limitation, we only manually collect the required concepts for advanced math. For other subjects, we use \texttt{GPT-4} to generate chapter names representing the required concepts. Each question is also categorized into chapters using \texttt{GPT-4}. More experimental details can be found in Appendix \ref{sec: Experiments Details of Calculating Concept Coverage Rate}.


\begin{table}[t!]
\centering

\resizebox{0.5\textwidth}{!}{\begin{tabular}{@{}lccccc@{}}

\toprule
Subject & Benchmark & min   & max  & mean& std \\ \midrule
\multirow{2}{*}{Professional Psychology} & C-EVAL & - & - & - & -\\
                   & CMMLU & 0.0 & 1.0 & 0.58 & 0.34 \\ \midrule
\multirow{2}{*}{Advanced Math} & C-EVAL  & 0.0      &   0.80   & 0.35     &  0.24 \\
                   & CMMLU &  0.0     &  0.75    & 0.41    &  0.18 \\ \bottomrule
\end{tabular}}

\caption{ Performance of \texttt{GPT-3.5-Turbo} on  subjects with manually collected required concepts across different chapter-level concepts.}
\label{tab: variance table}
\end{table}

\begin{table}[t!]
\centering
\resizebox{0.5\textwidth}{!}{
\begin{tabular}{@{}lcc@{}}
\toprule
Subject & Baichuan2-13B & Qwen1.5-MoE-A2.7B \\ \midrule
Computer Architecture & 0.50±0.17 & 0.72±0.14 \\
Computer Network & 0.59±0.19 & 0.80±0.17 \\
High School Biology & 0.52±0.25 & 0.68±0.32 \\
High School Geography & 0.73±0.16 & 0.84±0.19 \\
Marxism & 0.88±0.10 & 0.96±0.03 \\
High School History & 0.59±0.33 & 0.80±0.29 \\
Logic & 0.46±0.36 & 0.60±0.34 \\ \bottomrule
\end{tabular}}
\caption{The mean and standard deviation of model performance on randomly sampled C-EVAL subjects across different chapters.}
\label{tab:various of other subjects}
\end{table}

\paragraph{The concept coverage rate is Low.} Although psychology is a vital domain for AI to learn human behavior, it is absent in C-EVAL, and its chapter-level concept coverage rate in CMMLU is only 59\%. Questions with such an extremely low concept coverage rate could potentially mislead developers. This low coverage rate, although partly attributable to the broad spectrum of subjects within psychology, is not exclusive to this field. Even in more focused domains like advanced mathematics, the coverage rate is disappointingly low. Other subjects in C-EVAL (as shown in Table \ref{tab:concept coverage rate of other subjects}), such as High School History and Marxism, fare slightly better with coverage rates around 70\%. However, the average coverage rate across all subjects is only 54\%. This figure might fall short of user expectations, considering these questions are used as a representation of a subject.

\subsection{Variations in Performance Across Different Concepts \label{sec: concepts_variance}}

A high variance in scores across concepts can make the average accuracy more misleading. To explore this, we measure the variance of different concepts within a subject.
\paragraph{Setup} Following the method in section \S \ref{sec: Analysis on the Concept Coverage}, we categorize questions into chapter-level concepts and calculate the chapter-level performance variance.

\paragraph{The performance across different chapters varies greatly.}  As demonstrated in Table \ref{tab: variance table} and \ref{tab:various of other subjects}, powerful models like GPT-3.5-Turbo \citep{gpt-3.5} exhibit a standard deviation exceeding 10\% across different chapters. We also evaluate open-source models such as Baichuan2-13B \citep{yang2023baichuan} and Qwen1.5-MoE-A2.7B \citep{bai2023qwen}. These models persist in showing a standard deviation of over 10\% on subjects sampled from diverse disciplines. This high variance indicates that the final accuracy of these models is greatly influenced by the question set sampled.

\section{\name}
\begin{table}[t!]
\centering
\footnotesize
\begin{tabular}{@{}p{0.22\textwidth}lcllc@{}}
\toprule
\textbf{Category}                   & \textbf{\# C} & \textbf{\# Q} & $\mathbf{L_Q}$ & $\mathbf{L_A}$ \\ \midrule
\multicolumn{5}{c}{\textit{In terms of subject}}                                                            \\ 
Clinical \& Counseling Psychology &            56                &           156        &  38.9  &  12.0                      \\
Psychology of Personality          &             91                &             318     &  36.7  & 11.1                       \\
Abnormal Psychology                &           89                  &            268        & 35.8 & 12.4                    \\
History of Psychology              &            126                 &            472       & 27.4 & 10.5                    \\
General Psychology                 &           183                  &               605    & 35.3 & 9.4                  \\
Psychometrics                      &         115                   &                 368   & 43.2 & 10.1             \\
Social Psychology                  &           169                  &            559       & 26.2 & 13.7                  \\
Management Psychology              &           88                  &           315         & 37.2 & 10.4                   \\
Psychological Statistics           &           99                  &              311      & 57.0 & 8.4                    \\
Experimental Psychology            &          141                   &              413     & 59.3 & 9.3                    \\
Developmental Psychology           &          159                   &               580    & 30.7 & 11.3                    \\
Educational Psychology             &            67                 &                208    & 42.9 & 11.5               \\ \midrule
\multicolumn{5}{c}{\textit{In terms of split}}                                                              \\ 
Dev                                & -        & 60   & - & -        \\
Valid                              & -       & 428       & - & -       \\
Test                              & -       & 4085   & - & -             \\ \midrule
Total                              &    1383        &      4573  & - & -   \textbf{}     \\ \bottomrule
\end{tabular}

\caption{ Statistics of \name. The column ``\#C'' indicates the number of concepts we have annotated for each subject, with each concepts generating 4 questions and filtered by professional psychological annotators. The number of questions obtained after the review process is displayed in the column ``\#Q''. $L_Q$ and $L_A$ is the average length of a question and answer separately.}
\label{tab: Statistics of psybench}
\end{table}
Our proposed \name~ seeks to fill the gap by offering a complete benchmark for evaluating the knowledge and reasoning ability of models in psychology. To the best of our knowledge, we are the first to explain the issue of concept bias in Chinese MMLU-style benchmarks and mitigate this problem by constructing question sets that cover necessary concepts.

\subsection{Overview}
In \name, we manually summarize 1383 concepts from the \NEEM, then prompting \texttt{GPT-4} to generate questions for each to ensure comprehensive coverage of all key concepts. This method sets \name~ apart from prior Chinese MMLU benchmarks like CMMLU and C-EVAL, which relied on questions from online sources or textbooks. Our choice to use \texttt{GPT-4} for question generation was driven by two main reasons: 1) To avoid copyright issues associated with using existing questions, facilitating broader industrial application; 2) Many concepts lacked associated multiple-choice questions, posing a challenge for creating new ones. To maintain question quality, we develop specific prompts based on the Professional Counselor Examination formats and had them reviewed by three professional psychologists. \name~ differs from CMMLU and C-EVAL in the following ways:

\begin{figure}[t!]
\centering
  \includegraphics[width= \linewidth]{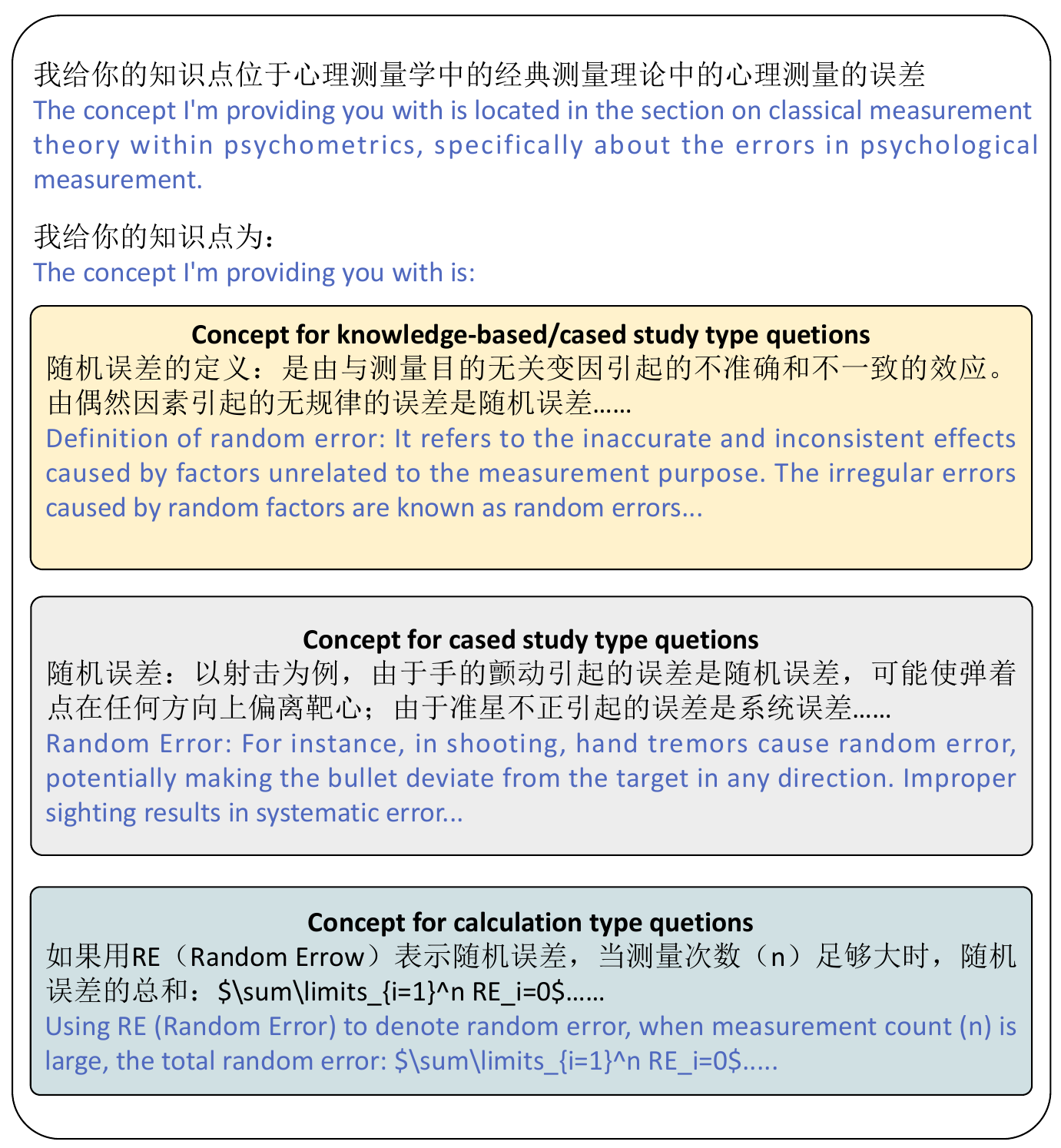}
  \caption{ An example of an annotator assigning a suitable prompt to a concept. For the concept ``random error'', we collect multiple descriptions. The appropriate prompt is assigned based on the type of description provided.}
  \label{fig:prompt_knowpoints}
\end{figure}

\begin{itemize}
\item \textbf{Low Concept Bias:} \name~ covers all required concepts to provide a low-biased results for users.
\item \textbf{Fine-grained Score:} Instead of only providing an average score, we also provide chapter-level scores to reflect fine-grained strengths and weaknesses for developers.
\item \textbf{An in-depth benchmark:}Instead of aiming to cover as many subjects as possible, we focus on providing the comprehensive and high in-depth coverage within the field of psychology, encompassing all college-level subjects.
\end{itemize}

\subsection{Data Collection}
\paragraph{Subject selection.}
As shown in Figure \ref{fig: intro_of_conceptpsy} and Table \ref{tab: Statistics of psybench}, We select  12 core standard subject within the discipline of psychology meticulously considered in accordance with both higher education standards and professional qualification requirements. We select 11 courses from the Peking University Core Curriculum Handbook for Undergraduate Program \footnote{\url{http://www.dean.pku.edu.cn/web/student_info.php?type=1&id=2}}. Additionally, we included Psychological Counselling as a fundamental subject within our data set to adhere to professional qualification criteria.

\paragraph{Concepts collection.}
We manually collect concepts based on the requirements of the 12 subjects outlined in \NEEM. This task is undertaken by eight graduate students over the course of one month. Each concept is summarized from the tutorials of the exams. An example of the collected concepts is shown in Figure \ref{fig:prompt_knowpoints}. For each concept, we summarize its description and manually determine which type of prompt (Calculation, Theory Understanding, Case Study) is suitable for generating questions about that concept (Prompts can be found in Appendix \ref{appendix:Prompts for Questions Generation}). To enhance \texttt{GPT-4}'s understanding of the provided concepts, we supply not only the concepts themselves but also the name of the subject, and the primary and secondary headings they fall under.

\paragraph{Prompts design.} In the process of designing prompts to generate high-quality questions, we conduct a thorough analysis of question types found in the Professional Counselor Examination, categorizing them into three distinct groups: (1) calculation; (2) theory understanding; (3) case study (Appendix \ref{appendix:Prompts for Questions Generation}). As depicted in Figure \ref{fig:prompt_knowpoints}, while collecting concepts, we manually assign one or more prompt types to each concept. This assignment strategy ensures the relevance and diversity of the generated questions. To maintain a relatively even number of questions for each concept, we prompt \texttt{GPT-4} to generate four questions per concept.

\begin{figure}[t!]
\centering
  \includegraphics[width=\linewidth]{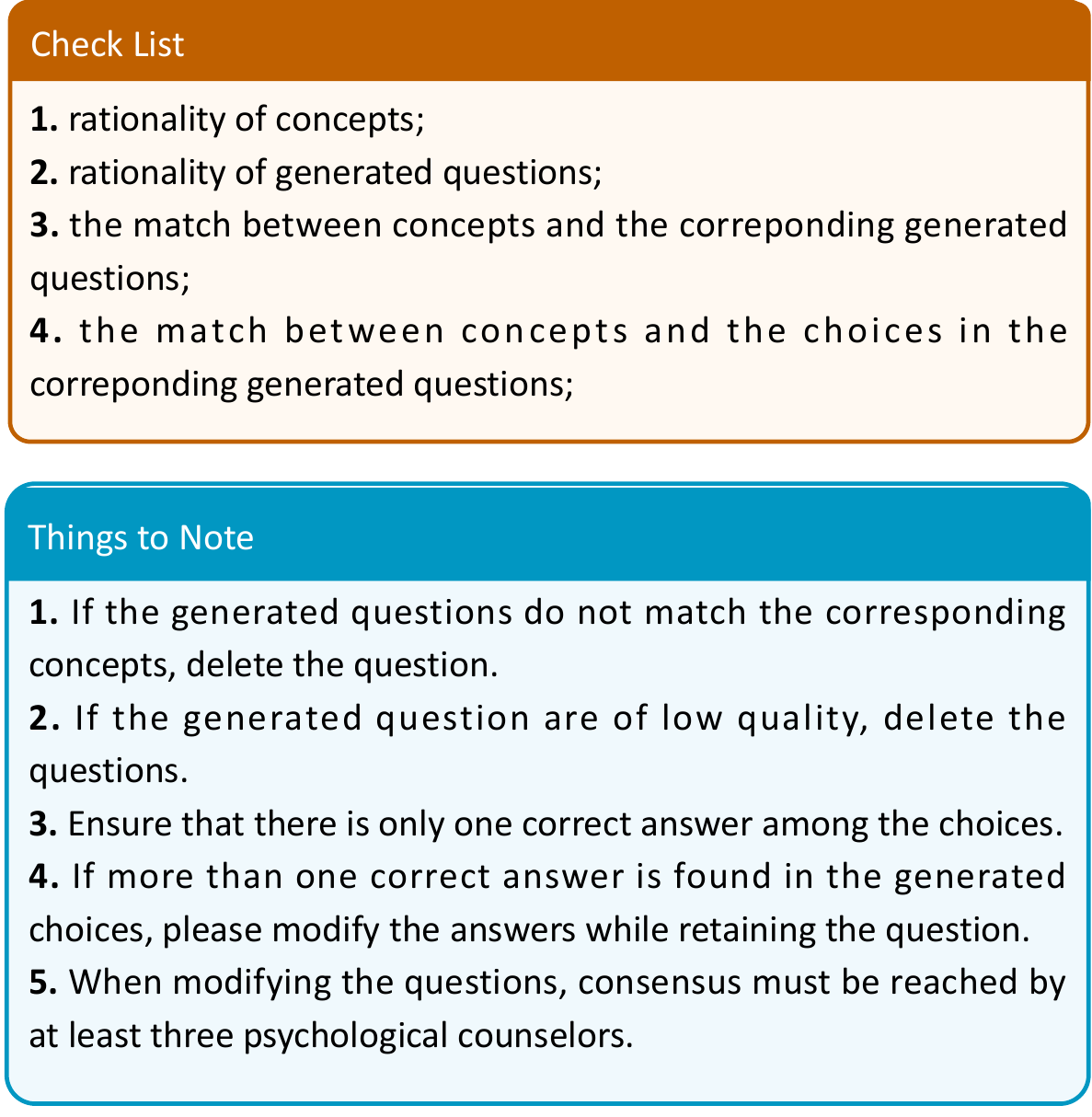}
  \caption{The review rules the question review process are as follows. Professional psychological annotators  will filter, modify, and review the generated questions based on these requirements.}
  \label{fig:check}
\end{figure}

\subsection{Questions Review}
To ensure the quality of our dataset, we employ three professional psychological annotators to meticulously review each question. Our main objective is to verify the accuracy of the questions and to remove any that are found to be unreasonable. Each question is subjected to a thorough evaluation by at least one professional psychologist, who identifies any aspects that are not reasonable. When inaccuracies are discovered, all professional psychologists collaborate to discuss and agree upon a suitable and precise correction strategy. The criteria for reviewing questions are detailed in Figure \ref{fig:check}. Throughout this process, the annotators check how relevant and accurate the knowledge points, the questions, and their connections are.
\section{Experiments}

\subsection{Dataset Statistics}
We carefully select 12 psychology subjects as shown in Table \ref{tab: Statistics of psybench}. For each concept, we generate 4 questions, leading to a total of 4573 high-quality questions after a thorough review and filtering process. Furthermore, as illustrated in Figure \ref{fig: intro_of_conceptpsy}, we collect chapter names based on the syllabus of these subjects and assign chapter-level labels to each generated question to provide detailed insights about the strengths and weaknesses.

\subsection{Setup}
We evaluate a diverse set of strong models to ensure a comprehensive evaluation. More details about the models can be found in Appendix \ref{sec: Details About Evaluated Models}. We conduct the evaluations using a 5-shot method. We set the temperature as 0, $top\_p$ as 1.0.

\begin{table*}[t!]
\centering
\setlength{\tabcolsep}{3pt} 
\resizebox{\textwidth}{!}{
\begin{tabular}{@{}p{0.25\textwidth}lccccccccccccc@{}}
\toprule
& \begin{tabular}[c]{@{}c@{}}Clinical \& \\ Counseling \end{tabular} 
& \begin{tabular}[c]{@{}c@{}}Psy of \\ Personality\end{tabular} 
& \begin{tabular}[c]{@{}c@{}}Abnormal \\ Psy\end{tabular} 
& \begin{tabular}[c]{@{}c@{}}History of \\ Psy\end{tabular} 
& \begin{tabular}[c]{@{}c@{}}General \\ Psy\end{tabular} 
& \begin{tabular}[c]{@{}c@{}} Psy- \\chometrics\end{tabular} 
& \begin{tabular}[c]{@{}c@{}}Social \\ Psy \end{tabular} 
& \begin{tabular}[c]{@{}c@{}}Management \\ Psy\end{tabular} 
& \begin{tabular}[c]{@{}c@{}}Psychological \\ Statistics\end{tabular} 
& \begin{tabular}[c]{@{}c@{}}Experimental \\ Psy\end{tabular} 
& \begin{tabular}[c]{@{}c@{}}Developmental \\ Psy\end{tabular} 
& \begin{tabular}[c]{@{}c@{}}Educational \\ Psy\end{tabular} 
& Avg \\
\midrule
\texttt{GPT-4} & 0.83 & \textbf{0.86} & 0.88 & \textbf{0.81} & 0.84 & \textbf{0.80} & \textbf{0.89} & 0.87 & \textbf{0.79} & \textbf{0.85} & 0.82 & \textbf{0.86} & 0.84 \\
\texttt{GPT-3.5-Turbo} & 0.78 & 0.74 & 0.80 & 0.67 & 0.73 & 0.70 & 0.82 & 0.80 & 0.64 & 0.74 & 0.77 & 0.81 & 0.75 \\
\texttt{Chi-Alpaca2-7B} & 0.59 & 0.58 & 0.59 & 0.52 & 0.45 & 0.51 & 0.63 & 0.62 & 0.46 & 0.58 & 0.59 & 0.67 & 0.57 \\
\texttt{Llama-2-13B-Chat} & 0.66 & 0.62 & 0.65 & 0.56 & 0.54 & 0.57 & 0.66 & 0.63 & 0.48 & 0.57 & 0.58 & 0.66 & 0.60 \\
\texttt{Chatglm2-6B} & 0.71 & 0.63 & 0.68 & 0.63 & 0.62 & 0.62 & 0.69 & 0.64 & 0.54 & 0.64 & 0.66 & 0.72 & 0.65 \\
\texttt{Llama-2-70B-Chat} & 0.74 & 0.66 & 0.73 & 0.62 & 0.61 & 0.59 & 0.75 & 0.70 & 0.52 & 0.63 & 0.69 & 0.69 & 0.66 \\
\texttt{Mistral-7B-Ins} & 0.73 & 0.65 & 0.66 & 0.59 & 0.62 & 0.63 & 0.71 & 0.72 & 0.52 & 0.65 & 0.66 & 0.74 & 0.66 \\
\texttt{Baichuan2-7B-Chat} & 0.72 & 0.69 & 0.77 & 0.65 & 0.67 & 0.66 & 0.75 & 0.72 & 0.53 & 0.66 & 0.70 & 0.76 & 0.69 \\
\texttt{Baichuan2-13B-Chat} & 0.78 & 0.73 & 0.80 & 0.68 & 0.73 & 0.69 & 0.82 & 0.75 & 0.60 & 0.71 & 0.74 & 0.83 & 0.74 \\

\texttt{Mixtral-8x7B-Ins} & 0.82 & 0.76 & 0.78 & 0.65 & 0.73 & 0.70 & 0.83 & 0.78 & 0.71 & 0.73 & 0.74 & 0.79 & 0.75 \\
\texttt{Qwen1.5-7B-Chat} & 0.83 & 0.75 & 0.84 & 0.66 & 0.75 & 0.67 & 0.79 & 0.81 & 0.61 & 0.72 & 0.73 & 0.83 & 0.75 \\
\texttt{Internlm2-7B-Chat} & 0.86 & 0.75 & 0.84 & 0.68 & 0.78 & 0.7 & 0.82 & 0.82 & 0.62 & 0.75 & 0.78 & 0.81 & 0.77 \\
\texttt{Yi-6B-Chat} & 0.85 & 0.78 & 0.89 & 0.71 & 0.84 & 0.70 & 0.85 & 0.86 & 0.62 & 0.72 & 0.78 & 0.83 & 0.79 \\
\texttt{Qwen1.5-72B-Chat} & 0.87 & 0.74 & 0.89 & 0.80 & 0.81 & 0.77 & 0.88 & 0.87 & 0.38 & 0.77 & 0.83 & 0.85 & 0.79 \\
\texttt{Qwen1.5-14B-Chat} & 0.86 & 0.80 & 0.85 & 0.72 & 0.83 & 0.72 & 0.83 & 0.85 & 0.67 & 0.81 & 0.79 & 0.85 & 0.80 \\
\texttt{Yi-34B-Chat} & \textbf{0.92} & 0.84 & \textbf{0.91} & \textbf{0.81} & \textbf{0.86} & 0.78 &0.88 & \textbf{0.90} & 0.69 & 0.84 & \textbf{0.86} & \textbf{0.86} & \textbf{0.85} \\

\bottomrule
\end{tabular}}
\caption{Performances on \name~ with different LLMs.}
\label{tab:main results}
\end{table*}

\begin{figure*}[t!] 
\centering 
\includegraphics[width=\linewidth]{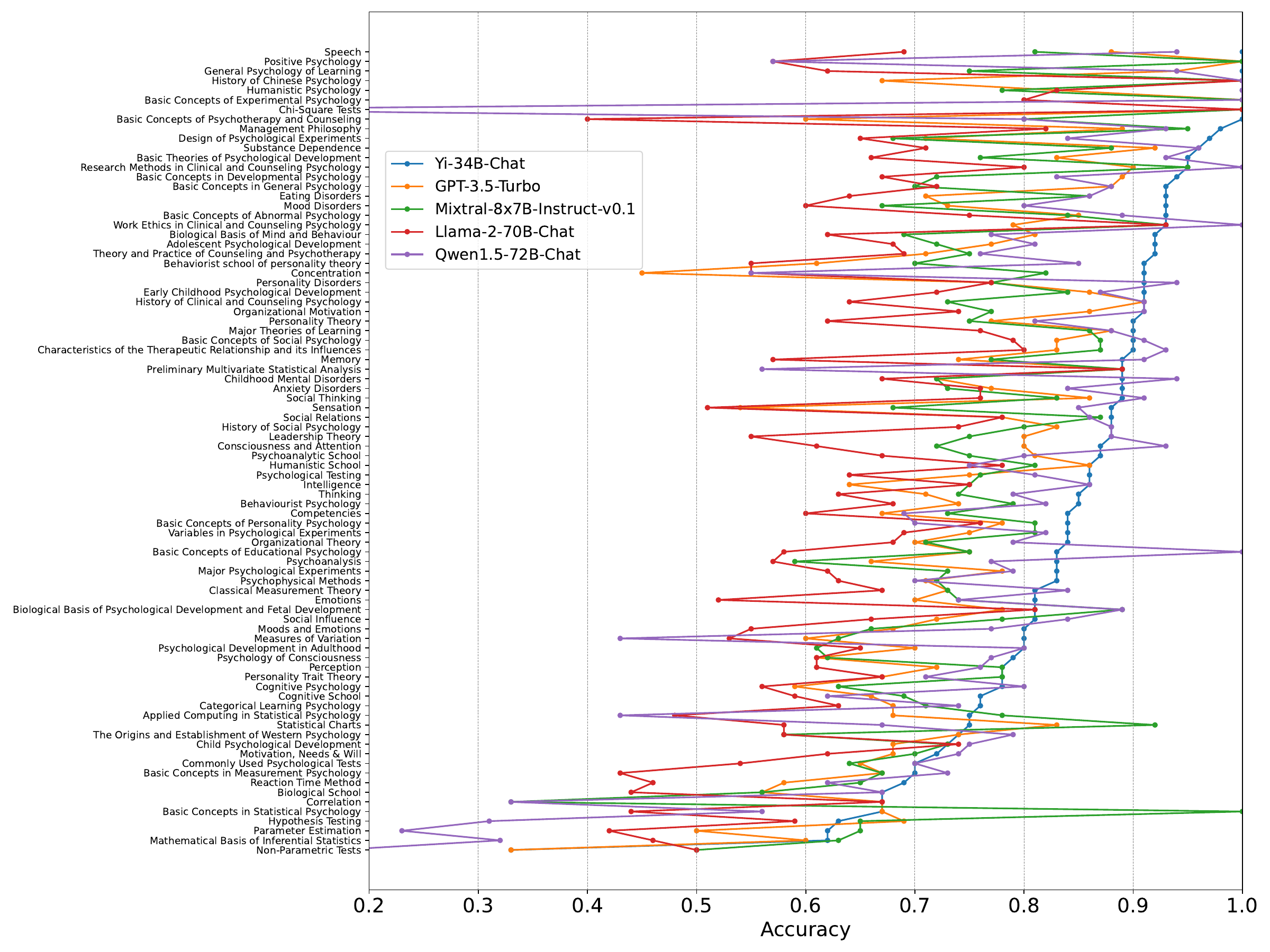} 
\caption{Concept-level results for models more than 34B, in addition to \texttt{GPT-3.5-Turbo}.}
\label{fig: conceptmap 72B} 
\end{figure*}
\subsection{Results}

The main results of different LLMs on \name~ are shown in Table \ref{tab:main results}. We also provide fine-grained results at Figure \ref{fig: conceptmap 72B}, \ref{fig: conceptmap 14B} and \ref{fig: conceptmap other}.

\textbf{Strong Chinese models achieve comparable overall results with \texttt{GPT-4}:} In Table \ref{tab:main results}, surprisingly, some Chinese models achieve comparable performance close to that of \texttt{GPT-4} on \name. Note that \texttt{GPT-4} have the concepts as input during the question generation process but not in the evaluation stage. Overall, \texttt{GPT-4} performed significantly better in subjects requiring more reasoning ability, such as Psychological Statistics (79\% compared to 69\%). However, in subjects requiring a deep understanding of the Chinese context and knowledge, such as Clinical \& Counseling, \texttt{Yi-34B-Chat} performed better.  This reflects the differences in abilities between the two models, indicating that GPT-4 might have insufficient training data in some areas of Chinese psychology, but its reasoning ability is much stronger. 

\textbf{Chinese models struggle with math and reasoning}: Analysis from Figure \ref{fig: conceptmap 72B}, \ref{fig: conceptmap 14B}, and \ref{fig: conceptmap other} reveals that Chinese models' performance drops on concepts requiring math and reasoning, like ``Mathematical Basis of Inferential Statistics'' and ``Non-Parametric Tests''. While these models excel in knowledge-based areas, they falter in math and reasoning segments. \texttt{Mixtral-8x7B-Ins} stands out by underperforming in knowledge questions but outshining others in math and reasoning. This suggests that despite high accuracy from training on extensive Chinese texts, Chinese models' math and reasoning skills, especially in psychology, need enhancement.

\textbf{Models show varied significant chapter performance:} There's notable variation in how models perform across chapters. \texttt{Yi-34B-Chat}, for example, despite leading in average accuracy, exhibits over 50\% performance differences between chapters—excelling in basic concepts but lagging in math reasoning, even behind \texttt{Mixtral-8x7B-Ins}  with its 10\% lower average accuracy. This underscores \name's role in offering detailed insights for model enhancement.

\textbf{Variability in conceptual mastery across models from the same series:} Although \texttt{Qwen1.5-14B-Chat} and \texttt{Qwen1.5-72B-Chat} achieve similar overall results, this does not mean their performance is similar across all chapters. As shown in Figures \ref{fig: conceptmap 72B} and \ref{fig: conceptmap 14B}, \texttt{Qwen1.5-72B-Chat} performs much worse in chapters that focus on math and reasoning, like "Non-Parametric Tests" and "Parameter Estimation." While the overall accuracy suggests that both models perform similarly, the detailed results from \name~ show that \texttt{Qwen1.5-72B-Chat} is generally better than \texttt{Qwen1.5-14B-Chat} in most chapters, except for a few math-related ones where it falls behind. This demonstrates how \name~ can highlight specific strengths and weaknesses, helping developers make targeted improvements.

\begin{table}[t!]
\centering
\resizebox{0.45\textwidth}{!}{\begin{tabular}{p{0.3\textwidth}lcc}
\toprule
Subjects & Pro & Non-Pro \\ 
\midrule
Clinical \& Counseling Psychology & 0.85 & 0.28 \\
Psychology of Personality & 0.82 & 0.24 \\
Abnormal Psychology & 0.89 & 0.32 \\
History of Psychology & 0.73 & 0.22 \\
General Psychology & 0.9 & 0.26 \\
Psychometrics & 0.75 & 0.3 \\
Social Psychology & 0.82 & 0.16 \\
Management Psychology & 0.8 & 0.36 \\
Psychological Statistic & 0.78 & 0.2 \\
Experimental Psychology & 0.82 & 0.32 \\
Developmental Psychology & 0.83 & 0.16 \\
Educational Psychology & 0.82 & 0.16 \\
\midrule
Avg & 0.82 & 0.25 \\
\bottomrule
\end{tabular}}
\caption{Human evaluation results of \name. ``Pro'' and ``Non-pro'' represent Professional Counselor and Non-Professional Student respectively.}
\label{tab:Human evaluation results}
\end{table}

\subsection{Ablation Study}
\textbf{Human baseline.} To assess \name's ability to distinguish between human expertise levels, we ask professional counselors and non-professional grad students to evaluate a subset of tasks. Table \ref{tab:Human evaluation results} shows significant performance differences (Appendix \ref{appendix: human evaluation}).

\textbf{The effect of linguist style.} 
To investigate the effect of different linguistic styles of generated questions, we paraphrase each question using \texttt{Qwen-70b-Chat} evaluate LLMs on it. The results are shown in Table \ref{tab:text_style_comparison}. There is a slight performance drop on paraphrased questions, potentially due to the lack of manual review for these paraphrased versions. Nonetheless, the consistency in model rankings and overall performance suggests that for multiple-choice questions, stylistic variations of the questions have limited effect.

\section{Related Works}
\textbf{Chinese MMLU Benchmarks:} While English language benchmarks continue to evolve~\citep{hendrycks2020measuring,huang2023ceval,li2023cmmlu,zhong2023agieval}, Chinese MMLU benchmarks remain underdeveloped. The CLUE benchmark \citep{xu-etal-2020-clue} is a widely used large-scale NLU benchmark for Chinese. AGIEval \citep{zhong2023agieval} expands this with questions from various Chinese exams, and MMCU \citep{zeng2023measuring} includes questions from diverse domains. The C-EVAL benchmark~\cite{huang2023ceval} gathers questions from different educational levels and professional exams, using non-paper-based and simulation questions to prevent dataset leakage. ConceptMath~\citep{wu2024conceptmath} evaluates LLMs in a concept-wise manner but focuses solely on mathematics, is not a comprehensive MMLU benchmark, and was released more than three months after our work.

\textbf{Benchmarks for LLMs in Psychology.} Although psychology is a crucial domain for achieving artificial intelligence, there are currently few benchmarks in this field. CMMLU has collected hundreds of questions as "professional psychology" to evaluate a model's knowledge understanding and reasoning capabilities. Other works~\citep{yang2023evaluations,amin2023will,lamichhane2023evaluation} approach this from a mental health perspective, utilizing various classification tasks, including various emotions, suicidal tendencies, and more. PsyEval~\citep{jin2023psyeval} further expands in the mental health field by adding tasks such as ``Diagnosis Prediction'', but lacking comprehensive subjects in college-level psychology.
\section{Discussion}

Domain benchmarks are crucial for adapting LLMs. However, a comprehensive benchmark for assessing models' knowledge and reasoning abilities in psychology is lacking. To address this gap, we introduce \name. In designing \name, we reveal concept bias issues in previous Chinese MMLU benchmarks and study concept coverage rates and performance variances. To tackle these issues, we propose a concept-wise question generation framework and provide chapter-level fine-grained results. Evaluation across various models demonstrates \name's effectiveness in highlighting model strengths and weaknesses.

\section*{Limitations}
We focused exclusively on concept bias within the Chinese MMLU dataset series, not exploring biases in other dimensions due to the labor-intensive nature of collecting the required concepts in a discipline, which necessitates trained annotators. Our experiments were thus confined to the field of psychology. Although our question generation method could be easily applied to other domains, due to copyright issues and the absence of corresponding multi-choice questions for many concepts, we opted to use GPT-4 for question generation instead of relying on human-designed questions.

\section*{Ethics Statement}
We have thoroughly examined our data to ensure that
there are no ethical issues. The data is generated by
GPT-4, using concepts prescribed by the National Entrance Examination for Postgraduates.
Furthermore, the generated data is subjected to rigorous scrutiny by professional psychologists to ensure its
ethical soundness.


\bibliography{anthology,custom}

\begin{thebibliography}{29}
\expandafter\ifx\csname natexlab\endcsname\relax\def\natexlab#1{#1}\fi

\bibitem[{Achiam et~al.(2023)Achiam, Adler, Agarwal, Ahmad, Akkaya, Aleman, Almeida, Altenschmidt, Altman, Anadkat et~al.}]{achiam2023gpt}
Josh Achiam, Steven Adler, Sandhini Agarwal, Lama Ahmad, Ilge Akkaya, Florencia~Leoni Aleman, Diogo Almeida, Janko Altenschmidt, Sam Altman, Shyamal Anadkat, et~al. 2023.
\newblock Gpt-4 technical report.
\newblock \emph{arXiv preprint arXiv:2303.08774}.

\bibitem[{AI et~al.(2024)AI, :, Young, Chen, Li, Huang, Zhang, Zhang, Li, Zhu, Chen, Chang, Yu, Liu, Liu, Yue, Yang, Yang, Yu, Xie, Huang, Hu, Ren, Niu, Nie, Xu, Liu, Wang, Cai, Gu, Liu, and Dai}]{ai2024yi}
01. AI, :, Alex Young, Bei Chen, Chao Li, Chengen Huang, Ge~Zhang, Guanwei Zhang, Heng Li, Jiangcheng Zhu, Jianqun Chen, Jing Chang, Kaidong Yu, Peng Liu, Qiang Liu, Shawn Yue, Senbin Yang, Shiming Yang, Tao Yu, Wen Xie, Wenhao Huang, Xiaohui Hu, Xiaoyi Ren, Xinyao Niu, Pengcheng Nie, Yuchi Xu, Yudong Liu, Yue Wang, Yuxuan Cai, Zhenyu Gu, Zhiyuan Liu, and Zonghong Dai. 2024.
\newblock \href {http://arxiv.org/abs/2403.04652} {Yi: Open foundation models by 01.ai}.

\bibitem[{Amin et~al.(2023)Amin, Cambria, and Schuller}]{amin2023will}
Mostafa~M Amin, Erik Cambria, and B~Schuller. 2023.
\newblock Will affective computing emerge from foundation models and general ai.
\newblock \emph{A first evaluation on ChatGPT. ArXiv, abs/2303.03186}.

\bibitem[{Anthropic(2023)}]{claude3}
Anthropic. 2023.
\newblock \href {https://anthropic.com/news/introducing-claude} {Introducing claude}.

\bibitem[{Bai et~al.(2023{\natexlab{a}})Bai, Bai, Chu, Cui, Dang, Deng, Fan, Ge, Han, Huang, Hui, Ji, Li, Lin, Lin, Liu, Liu, Lu, Lu, Ma, Men, Ren, Ren, Tan, Tan, Tu, Wang, Wang, Wang, Wu, Xu, Xu, Yang, Yang, Yang, Yang, Yao, Yu, Yuan, Yuan, Zhang, Zhang, Zhang, Zhang, Zhou, Zhou, Zhou, and Zhu}]{qwen}
Jinze Bai, Shuai Bai, Yunfei Chu, Zeyu Cui, Kai Dang, Xiaodong Deng, Yang Fan, Wenbin Ge, Yu~Han, Fei Huang, Binyuan Hui, Luo Ji, Mei Li, Junyang Lin, Runji Lin, Dayiheng Liu, Gao Liu, Chengqiang Lu, Keming Lu, Jianxin Ma, Rui Men, Xingzhang Ren, Xuancheng Ren, Chuanqi Tan, Sinan Tan, Jianhong Tu, Peng Wang, Shijie Wang, Wei Wang, Shengguang Wu, Benfeng Xu, Jin Xu, An~Yang, Hao Yang, Jian Yang, Shusheng Yang, Yang Yao, Bowen Yu, Hongyi Yuan, Zheng Yuan, Jianwei Zhang, Xingxuan Zhang, Yichang Zhang, Zhenru Zhang, Chang Zhou, Jingren Zhou, Xiaohuan Zhou, and Tianhang Zhu. 2023{\natexlab{a}}.
\newblock Qwen technical report.
\newblock \emph{arXiv preprint arXiv:2309.16609}.

\bibitem[{Bai et~al.(2023{\natexlab{b}})Bai, Bai, Chu, Cui, Dang, Deng, Fan, Ge, Han, Huang et~al.}]{bai2023qwen}
Jinze Bai, Shuai Bai, Yunfei Chu, Zeyu Cui, Kai Dang, Xiaodong Deng, Yang Fan, Wenbin Ge, Yu~Han, Fei Huang, et~al. 2023{\natexlab{b}}.
\newblock Qwen technical report.
\newblock \emph{arXiv preprint arXiv:2309.16609}.

\bibitem[{Cai et~al.(2024)Cai, Cao, Chen, Chen, Chen, Chen, Chen, Chen, Chen, Chu, Dong, Duan, Fan, Fei, Gao, Ge, Gu, Gu, Gui, Guo, Guo, He, Hu, Huang, Jiang, Jiao, Jin, Lei, Li, Li, Li, Li, Li, Li, Liu, Liu, Hong, Liu, Liu, Liu, Lv, Lv, Lv, Ma, Ma, Ma, Ning, Ouyang, Qiu, Qu, Shang, Shao, Song, Song, Sui, Sun, Sun, Tang, Wang, Wang, Wang, Wang, Wang, Wang, Wang, Wei, Weng, Wu, Xiong, Xu, Xu, Yan, Yan, Yang, Ye, Ying, Yu, Yu, Zang, Zhang, Zhang, Zhang, Zhang, Zhang, Zhang, Zhang, Zhang, Zhang, Zhang, Zhang, Zhao, Zhao, Zhao, Zhou, Zhou, Zhuo, Zou, Qiu, Qiao, and Lin}]{cai2024internlm2}
Zheng Cai, Maosong Cao, Haojiong Chen, Kai Chen, Keyu Chen, Xin Chen, Xun Chen, Zehui Chen, Zhi Chen, Pei Chu, Xiaoyi Dong, Haodong Duan, Qi~Fan, Zhaoye Fei, Yang Gao, Jiaye Ge, Chenya Gu, Yuzhe Gu, Tao Gui, Aijia Guo, Qipeng Guo, Conghui He, Yingfan Hu, Ting Huang, Tao Jiang, Penglong Jiao, Zhenjiang Jin, Zhikai Lei, Jiaxing Li, Jingwen Li, Linyang Li, Shuaibin Li, Wei Li, Yining Li, Hongwei Liu, Jiangning Liu, Jiawei Hong, Kaiwen Liu, Kuikun Liu, Xiaoran Liu, Chengqi Lv, Haijun Lv, Kai Lv, Li~Ma, Runyuan Ma, Zerun Ma, Wenchang Ning, Linke Ouyang, Jiantao Qiu, Yuan Qu, Fukai Shang, Yunfan Shao, Demin Song, Zifan Song, Zhihao Sui, Peng Sun, Yu~Sun, Huanze Tang, Bin Wang, Guoteng Wang, Jiaqi Wang, Jiayu Wang, Rui Wang, Yudong Wang, Ziyi Wang, Xingjian Wei, Qizhen Weng, Fan Wu, Yingtong Xiong, Chao Xu, Ruiliang Xu, Hang Yan, Yirong Yan, Xiaogui Yang, Haochen Ye, Huaiyuan Ying, Jia Yu, Jing Yu, Yuhang Zang, Chuyu Zhang, Li~Zhang, Pan Zhang, Peng Zhang, Ruijie Zhang, Shuo Zhang, Songyang Zhang, Wenjian Zhang,
  Wenwei Zhang, Xingcheng Zhang, Xinyue Zhang, Hui Zhao, Qian Zhao, Xiaomeng Zhao, Fengzhe Zhou, Zaida Zhou, Jingming Zhuo, Yicheng Zou, Xipeng Qiu, Yu~Qiao, and Dahua Lin. 2024.
\newblock \href {http://arxiv.org/abs/2403.17297} {Internlm2 technical report}.

\bibitem[{Chen et~al.(2024)Chen, Fang, Singla, and Dredze}]{chen2024benchmarking}
Hanjie Chen, Zhouxiang Fang, Yash Singla, and Mark Dredze. 2024.
\newblock Benchmarking large language models on answering and explaining challenging medical questions.
\newblock \emph{arXiv preprint arXiv:2402.18060}.

\bibitem[{Cui et~al.(2023)Cui, Yang, and Yao}]{Chinese-LLaMA-Alpaca}
Yiming Cui, Ziqing Yang, and Xin Yao. 2023.
\newblock \href {https://arxiv.org/abs/2304.08177} {Efficient and effective text encoding for chinese llama and alpaca}.
\newblock \emph{arXiv preprint arXiv:2304.08177}.

\bibitem[{DeepSeek-AI(2024)}]{deepseekv2}
DeepSeek-AI. 2024.
\newblock \href {http://arxiv.org/abs/2405.04434} {Deepseek-v2: A strong, economical, and efficient mixture-of-experts language model}.

\bibitem[{Du et~al.(2022)Du, Qian, Liu, Ding, Qiu, Yang, and Tang}]{du2022glm}
Zhengxiao Du, Yujie Qian, Xiao Liu, Ming Ding, Jiezhong Qiu, Zhilin Yang, and Jie Tang. 2022.
\newblock Glm: General language model pretraining with autoregressive blank infilling.
\newblock In \emph{Proceedings of the 60th Annual Meeting of the Association for Computational Linguistics (Volume 1: Long Papers)}, pages 320--335.

\bibitem[{Fei et~al.(2023)Fei, Shen, Zhu, Zhou, Han, Zhang, Chen, Shen, and Ge}]{fei2023lawbench}
Zhiwei Fei, Xiaoyu Shen, Dawei Zhu, Fengzhe Zhou, Zhuo Han, Songyang Zhang, Kai Chen, Zongwen Shen, and Jidong Ge. 2023.
\newblock Lawbench: Benchmarking legal knowledge of large language models.
\newblock \emph{arXiv preprint arXiv:2309.16289}.

\bibitem[{Hendrycks et~al.(2020)Hendrycks, Burns, Basart, Zou, Mazeika, Song, and Steinhardt}]{hendrycks2020measuring}
Dan Hendrycks, Collin Burns, Steven Basart, Andy Zou, Mantas Mazeika, Dawn Song, and Jacob Steinhardt. 2020.
\newblock Measuring massive multitask language understanding.
\newblock \emph{arXiv preprint arXiv:2009.03300}.

\bibitem[{Huang et~al.(2023)Huang, Bai, Zhu, Zhang, Zhang, Su, Liu, Lv, Zhang, Lei, Fu, Sun, and He}]{huang2023ceval}
Yuzhen Huang, Yuzhuo Bai, Zhihao Zhu, Junlei Zhang, Jinghan Zhang, Tangjun Su, Junteng Liu, Chuancheng Lv, Yikai Zhang, Jiayi Lei, Yao Fu, Maosong Sun, and Junxian He. 2023.
\newblock C-eval: A multi-level multi-discipline chinese evaluation suite for foundation models.
\newblock In \emph{Advances in Neural Information Processing Systems}.

\bibitem[{Jiang et~al.(2023)Jiang, Sablayrolles, Mensch, Bamford, Chaplot, Casas, Bressand, Lengyel, Lample, Saulnier et~al.}]{jiang2023mistral}
Albert~Q Jiang, Alexandre Sablayrolles, Arthur Mensch, Chris Bamford, Devendra~Singh Chaplot, Diego de~las Casas, Florian Bressand, Gianna Lengyel, Guillaume Lample, Lucile Saulnier, et~al. 2023.
\newblock Mistral 7b.
\newblock \emph{arXiv preprint arXiv:2310.06825}.

\bibitem[{Jiang et~al.(2024)Jiang, Sablayrolles, Roux, Mensch, Savary, Bamford, Chaplot, Casas, Hanna, Bressand et~al.}]{jiang2024mixtral}
Albert~Q Jiang, Alexandre Sablayrolles, Antoine Roux, Arthur Mensch, Blanche Savary, Chris Bamford, Devendra~Singh Chaplot, Diego de~las Casas, Emma~Bou Hanna, Florian Bressand, et~al. 2024.
\newblock Mixtral of experts.
\newblock \emph{arXiv preprint arXiv:2401.04088}.

\bibitem[{Jin et~al.(2023)Jin, Chen, Wu, and Zhu}]{jin2023psyeval}
Haoan Jin, Siyuan Chen, Mengyue Wu, and Kenny~Q Zhu. 2023.
\newblock Psyeval: A comprehensive large language model evaluation benchmark for mental health.
\newblock \emph{arXiv preprint arXiv:2311.09189}.

\bibitem[{Kwon et~al.(2023)Kwon, Li, Zhuang, Sheng, Zheng, Yu, Gonzalez, Zhang, and Stoica}]{kwon2023efficient}
Woosuk Kwon, Zhuohan Li, Siyuan Zhuang, Ying Sheng, Lianmin Zheng, Cody~Hao Yu, Joseph~E. Gonzalez, Hao Zhang, and Ion Stoica. 2023.
\newblock Efficient memory management for large language model serving with pagedattention.
\newblock In \emph{Proceedings of the ACM SIGOPS 29th Symposium on Operating Systems Principles}.

\bibitem[{Lamichhane(2023)}]{lamichhane2023evaluation}
Bishal Lamichhane. 2023.
\newblock Evaluation of chatgpt for nlp-based mental health applications.
\newblock \emph{arXiv preprint arXiv:2303.15727}.

\bibitem[{Li et~al.(2023)Li, Zhang, Koto, Yang, Zhao, Gong, Duan, and Baldwin}]{li2023cmmlu}
Haonan Li, Yixuan Zhang, Fajri Koto, Yifei Yang, Hai Zhao, Yeyun Gong, Nan Duan, and Timothy Baldwin. 2023.
\newblock Cmmlu: Measuring massive multitask language understanding in chinese.
\newblock \emph{arXiv preprint arXiv:2306.09212}.

\bibitem[{OpenAI(2022)}]{gpt-3.5}
OpenAI. 2022.
\newblock \href {https://openai.com/blog/chatgpt} {Introducing chatgpt}.

\bibitem[{Touvron et~al.(2023)Touvron, Martin, Stone, Albert, Almahairi, Babaei, Bashlykov, Batra, Bhargava, Bhosale et~al.}]{touvron2023llama}
Hugo Touvron, Louis Martin, Kevin Stone, Peter Albert, Amjad Almahairi, Yasmine Babaei, Nikolay Bashlykov, Soumya Batra, Prajjwal Bhargava, Shruti Bhosale, et~al. 2023.
\newblock Llama 2: Open foundation and fine-tuned chat models.
\newblock \emph{arXiv preprint arXiv:2307.09288}.

\bibitem[{Wu et~al.(2024)Wu, Liu, Bu, Liu, Zhou, Zhang, Zhang, Bai, Chen, Ge et~al.}]{wu2024conceptmath}
Yanan Wu, Jie Liu, Xingyuan Bu, Jiaheng Liu, Zhanhui Zhou, Yuanxing Zhang, Chenchen Zhang, Zhiqi Bai, Haibin Chen, Tiezheng Ge, et~al. 2024.
\newblock Conceptmath: A bilingual concept-wise benchmark for measuring mathematical reasoning of large language models.
\newblock \emph{arXiv preprint arXiv:2402.14660}.

\bibitem[{Xu et~al.(2020)Xu, Hu, Zhang, Li, Cao, Li, Xu, Sun, Yu, Yu, Tian, Dong, Liu, Shi, Cui, Li, Zeng, Wang, Xie, Li, Patterson, Tian, Zhang, Zhou, Liu, Zhao, Zhao, Yue, Zhang, Yang, Richardson, and Lan}]{xu-etal-2020-clue}
Liang Xu, Hai Hu, Xuanwei Zhang, Lu~Li, Chenjie Cao, Yudong Li, Yechen Xu, Kai Sun, Dian Yu, Cong Yu, Yin Tian, Qianqian Dong, Weitang Liu, Bo~Shi, Yiming Cui, Junyi Li, Jun Zeng, Rongzhao Wang, Weijian Xie, Yanting Li, Yina Patterson, Zuoyu Tian, Yiwen Zhang, He~Zhou, Shaoweihua Liu, Zhe Zhao, Qipeng Zhao, Cong Yue, Xinrui Zhang, Zhengliang Yang, Kyle Richardson, and Zhenzhong Lan. 2020.
\newblock \href {https://doi.org/10.18653/v1/2020.coling-main.419} {{CLUE}: A {C}hinese language understanding evaluation benchmark}.
\newblock In \emph{Proceedings of the 28th International Conference on Computational Linguistics}, pages 4762--4772, Barcelona, Spain (Online). International Committee on Computational Linguistics.

\bibitem[{Yang et~al.(2023{\natexlab{a}})Yang, Xiao, Wang, Zhang, Bian, Yin, Lv, Pan, Wang, Yan et~al.}]{yang2023baichuan}
Aiyuan Yang, Bin Xiao, Bingning Wang, Borong Zhang, Ce~Bian, Chao Yin, Chenxu Lv, Da~Pan, Dian Wang, Dong Yan, et~al. 2023{\natexlab{a}}.
\newblock Baichuan 2: Open large-scale language models.
\newblock \emph{arXiv preprint arXiv:2309.10305}.

\bibitem[{Yang et~al.(2023{\natexlab{b}})Yang, Ji, Zhang, Xie, and Ananiadou}]{yang2023evaluations}
Kailai Yang, Shaoxiong Ji, Tianlin Zhang, Qianqian Xie, and Sophia Ananiadou. 2023{\natexlab{b}}.
\newblock On the evaluations of chatgpt and emotion-enhanced prompting for mental health analysis.
\newblock \emph{arXiv preprint arXiv:2304.03347}.

\bibitem[{Zeng(2023)}]{zeng2023measuring}
Hui Zeng. 2023.
\newblock Measuring massive multitask chinese understanding.
\newblock \emph{arXiv preprint arXiv:2304.12986}.

\bibitem[{Zhang et~al.(2023)Zhang, Cai, Liu, Yang, Dai, Liao, Qin, Li, Liu, Liu et~al.}]{zhang2023fineval}
Liwen Zhang, Weige Cai, Zhaowei Liu, Zhi Yang, Wei Dai, Yujie Liao, Qianru Qin, Yifei Li, Xingyu Liu, Zhiqiang Liu, et~al. 2023.
\newblock Fineval: A chinese financial domain knowledge evaluation benchmark for large language models.
\newblock \emph{arXiv preprint arXiv:2308.09975}.

\bibitem[{Zhong et~al.(2023)Zhong, Cui, Guo, Liang, Lu, Wang, Saied, Chen, and Duan}]{zhong2023agieval}
Wanjun Zhong, Ruixiang Cui, Yiduo Guo, Yaobo Liang, Shuai Lu, Yanlin Wang, Amin Saied, Weizhu Chen, and Nan Duan. 2023.
\newblock Agieval: A human-centric benchmark for evaluating foundation models.
\newblock \emph{arXiv preprint arXiv:2304.06364}.

\end{thebibliography}
\bibliographystyle{acl_natbib}
\clearpage
\newpage
\appendix
\section{Human Evaluation \label{appendix: human evaluation}}

We also provide human performance on the subset of \name. Specifically, we randomly sample 50 questions from each of 12 subjects, totaling 600 questions. A professional counselor independently completes all 600 questions, while ten non-experts are each assigned different subjects for evaluation. The results are shown in Table \ref{tab:Human evaluation results}.  The professional counselor significantly outperforms GPT-3.5, achieving an average score of 82\%. However, the performance of non-expert undergraduates and graduate students is poor, averaging 25\%, which is close to random. This underscores our benchmark's demand for specialized knowledge in psychology and its effectiveness in distinguishing whether a model has mastered these professional concepts.
\section{The effect of Linguistic Style of generated questions}

To investigate the effect of linguistic style, we paraphrase each question using \texttt{Qwen-70b-Chat}. We then evaluated the performance difference between its style and \texttt{GPT-4}'s style. The results are summarized in the table below:

\begin{table*}[t!]
\centering
\setlength{\tabcolsep}{3pt} 
\resizebox{\textwidth}{!}{
\begin{tabular}{@{}p{0.25\textwidth}lccccccccccccc@{}}
\toprule
& \begin{tabular}[c]{@{}c@{}}Clinical \& \\ Counseling \end{tabular} 
& \begin{tabular}[c]{@{}c@{}}Psy of \\ Personality\end{tabular} 
& \begin{tabular}[c]{@{}c@{}}Abnormal \\ Psy\end{tabular} 
& \begin{tabular}[c]{@{}c@{}}History of \\ Psy\end{tabular} 
& \begin{tabular}[c]{@{}c@{}}General \\ Psy\end{tabular} 
& \begin{tabular}[c]{@{}c@{}} Psy- \\chometrics\end{tabular} 
& \begin{tabular}[c]{@{}c@{}}Social \\ Psy \end{tabular} 
& \begin{tabular}[c]{@{}c@{}}Management \\ Psy\end{tabular} 
& \begin{tabular}[c]{@{}c@{}}Psychological \\ Statistics\end{tabular} 
& \begin{tabular}[c]{@{}c@{}}Experimental \\ Psy\end{tabular} 
& \begin{tabular}[c]{@{}c@{}}Developmental \\ Psy\end{tabular} 
& \begin{tabular}[c]{@{}c@{}}Educational \\ Psy\end{tabular} 
& Avg \\
\midrule
Mistral-7B-Instruct & 0.71 & 0.63 & 0.64 & 0.58 & 0.57 & 0.59 & 0.69 & 0.71 & 0.53 & 0.64 & 0.62 & 0.69 & 0.63 \\
gpt-3.5-Turbo       & 0.80 & 0.72 & 0.79 & 0.67 & 0.70 & 0.68 & 0.77 & 0.78 & 0.60 & 0.73 & 0.71 & 0.76 & 0.73 \\
Yi-6B-Chat          & 0.80 & 0.75 & 0.87 & 0.67 & 0.80 & 0.70 & 0.83 & 0.83 & 0.58 & 0.69 & 0.76 & 0.86 & 0.76 \\
Yi-34B-Chat         & 0.86 & 0.80 & 0.90 & 0.78 & 0.83 & 0.78 & 0.87 & 0.88 & 0.70 & 0.82 & 0.82 & 0.85 & 0.82 \\
\bottomrule
\end{tabular}}
\caption{Performances of models on questions paraphrased by \texttt{Qwen-70b-Chat}}
\label{tab:text_style_comparison}
\end{table*}

In comparison to Table \ref{tab:main results}, models perform slightly worse on paraphrased questions generated by \texttt{Qwen-70b-Chat}. This may be because the paraphrased questions are not manually reviewed. However, despite the style changes, the model rankings and overall performance remained consistent. This indicates that for multiple-choice questions, the style of the questions has minimal impact on evaluation.
\begin{table}[t!]
\centering
\resizebox{0.5\textwidth}{!}{\begin{tabular}{lcc}
\toprule
Subjects & Win Rate & Tie Rate \\ 
\midrule
Clinical \& Counseling Psychology & 0 & 0.8 \\
Psychology of Personality & 0 & 0.9 \\
Abnormal Psychology & 0 & 0.8 \\
History of Psychology & 0.1 & 0.8 \\
General Psychology & 0.1 & 0.9 \\
Psychometrics & 0.1 & 0.7 \\
Social Psychology & 0.82 & 0.16 \\
Management Psychology & 0.1 & 0.8 \\
Psychological Statistic & 0.4 & 0.5 \\
Experimental Psychology & 0.2 & 0.7 \\
Developmental Psychology & 0.4 & 0.4 \\
Educational Psychology & 0.3 & 0.7 \\
\midrule
Avg & 0.27 & 0.62 \\
\bottomrule
\end{tabular}}
\caption{We hire two professional counselor to compete the quality of generated questions and human-designed questions. }
\label{tab:quality}
\end{table}

\section{Competition with Human-designed Questions}
We randomly selected 15-20 questions for each subject and an equivalent number of questions from real exams available online. We shuffle the order of the questions, presenting one generated and one real question side by side, and ask two psychological annotators to judge which was better or if there was a tie. The averaged results are shown in Table \ref{tab:quality}. The ``Win Rate'' indicates the proportion of instances where our generated questions were rated better than the real ones. As can be seen, the majority of the generated questions matched or even surpassed the performance of actual exam questions. This success can be attributed to two factors: 1) our prompts are a carefully designed summary of the question types found in Chinese psychological professional examinations; 2) GPT-4's robust capabilities in synthetic data.

\section{Chapter-level Statistic of ConceptPsy}
\label{sec: Chapter-level Statistic of ConceptPsy}

In Table \ref{tab:chapter-level-1} and Table \ref{tab:chapter-level-2}, we present the chapters of each subject along with the corresponding number of concepts and questions.

\begin{table*}[t]
\centering
\resizebox{\textwidth}{!}{
\begin{tabular}{@{}lccc@{}}
\toprule
\midrule
Subject                                            & Chapter                                                             & \# C & \# Q \\ \midrule
\multirow{6}{*}{Clinical \& Counseling Psychology} & History of Clinical and Counseling Psychology                        & 4    & 13   \\
                                                   & Basic concepts of psychotherapy and counseling                  & 3    & 7    \\
                                                   & Characteristics of the therapeutic relationship and its influences          & 10   & 32   \\
                                                   & Work Ethics in Clinical and Counseling Psychology                    & 5    & 16   \\
                                                   & Research Methods in Clinical and Counseling Psychology               & 8    & 25   \\
                                                   & Theory and Practice of Counseling and Psychotherapy                & 25   & 63   \\ \midrule
\multirow{8}{*}{Psychology of Personality}         & Basic Concepts of Personality Psychology                            & 12   & 43   \\
                                                   & Psychoanalytic School                                              & 23   & 84   \\
                                                   & Behaviorist school of personality theory                                   & 10   & 37   \\
                                                   & Cognitive School                                                    & 9    & 33   \\
                                                   & Humanistic School                                                   & 11   & 40   \\
                                                   & Personality Trait Theory                                        & 18   & 57   \\
                                                   & Biological School                                                   & 4    & 13   \\
                                                   & Positive Psychology                                                 & 4    & 11   \\ \midrule
\multirow{7}{*}{Abnormal Psychology}               & Basic Concepts of Abnormal Psychology                               & 17   & 60   \\
                                                   & Anxiety Disorders                                                   & 26   & 81   \\
                                                   & Mood Disorders                                                      & 7    & 19   \\
                                                   & Eating Disorders                                                    & 6    & 18   \\
                                                   & Personality Disorders                                               & 14   & 40   \\
                                                   & Substance Dependence                                                & 10   & 28   \\
                                                   & Childhood Mental Disorders                                   & 8    & 22   \\ \midrule
\multirow{7}{*}{History of Psychology}             & The Origins and Establishment of Western Psychology                 & 17   & 62   \\
                                                   & Psychology of Consciousness                                         & 40   & 154  \\
                                                   & Behaviourist Psychology                                             & 21   & 82   \\
                                                   & Psychoanalysis                                                      & 25   & 93   \\
                                                   & Cognitive Psychology                                                & 13   & 48   \\
                                                   & Humanistic Psychology                                               & 8    & 30   \\
                                                   & History of Chinese Psychology                                       & 1    & 3    \\ \midrule
\multirow{12}{*}{General Psychology}               & Basic Concepts in General Psychology                                & 15   & 48   \\
                                                   & Biological Basis of Mind and Behaviour                              & 9    & 33   \\
                                                   & Consciousness and Attention                                         & 15   & 51   \\
                                                   & Sensation                                                           & 15   & 46   \\
                                                   & Perception                                                          & 17   & 60   \\
                                                   & Memory                                                              & 16   & 53   \\
                                                   & Thinking                                                            & 21   & 67   \\
                                                   & Speech                                                              & 7    & 21   \\
                                                   & Moods and Emotions                                                  & 15   & 50   \\
                                                   & Motivation, Needs \& Will                                      & 15   & 52   \\
                                                   & Competencies                                                          & 15   & 50   \\
                                                   & Personality Theory                                                  & 22   & 74   \\ \midrule
\multirow{4}{*}{Psychometrics}                     & Basic Concepts in Measurement Psychology                            & 12   & 41   \\
                                                   & Classical Measurement Theory                                        & 24   & 91   \\
                                                   & Basic Concepts of Psychological Testing                                                 & 40   & 105  \\
                                                   & Commonly Used Psychological Tests                                          & 39   & 129  \\ \midrule \bottomrule

\end{tabular}
}
\caption{Details of the number of Concepts and Questions in each Chapter.}
\label{tab:chapter-level-1}
\end{table*}

\begin{table*}[t]
\centering
\resizebox{\textwidth}{!}{
\begin{tabular}{@{}lccc@{}}
\toprule
\midrule
Subject                                            & Chapter                                                             & \# C & \# Q \\ \midrule

\multirow{5}{*}{Social Psychology}                 & History of Social Psychology                                        & 40   & 134  \\
                                                   & Social Thinking                                                     & 48   & 163  \\
                                                   & Social Relations                                                & 27   & 94   \\
                                                   & Social Influence                                                    & 24   & 79   \\
                                                   & Basic Concepts of Social Psychology                                 & 30   & 89   \\ \midrule
\multirow{4}{*}{Management Psychology}             & Management Philosophy                                            & 18   & 65   \\
                                                   & Organizational Motivation                                           & 18   & 80   \\
                                                   & Leadership Theory                                                   & 18   & 65   \\
                                                   & Organizational Theory                                               & 33   & 105  \\ \midrule
\multirow{12}{*}{Psychological Statistics}         & Basic Concepts in Statistical Psychology                            & 3    & 12   \\
                                                   & Statistical Charts                                                  & 5    & 15   \\
                                                   & Concentration                                              & 3    & 13   \\
                                                   & Measures of Variation                                                 & 7    & 34   \\
                                                   & Correlation                                                & 2    & 5    \\
                                                   & Mathematical Basis of Inferential Statistics                  & 16   & 65   \\
                                                   & Parameter Estimation                                                & 13   & 28   \\
                                                   & Hypothesis Testing                                                  & 13   & 53   \\
                                                   & Applied Computing in Statistical Psychology                      & 21   & 62   \\
                                                   & Chi-Square Tests                                                     & 2    & 5    \\
                                                   & Non-Parametric Tests                                                 & 2    & 8    \\
                                                   & Preliminary Multivariate Statistical Analysis                       & 2    & 11   \\ \midrule
\multirow{6}{*}{Experimental Psychology}           & Basic Concepts of Experimental Psychology                           & 3    & 12   \\
                                                   & Variables in Psychological Experiments                              & 23   & 75   \\
                                                   & Design of Psychological Experiments                                 & 12   & 38   \\
                                                   & Reaction Time Method                                                & 9    & 33   \\
                                                   & Psychophysical Methods                                               & 40   & 97   \\
                                                   & Major Psychological Experiments                                     & 47   & 158  \\ \midrule
\multirow{9}{*}{Developmental Psychology}          & Basic Concepts in Developmental Psychology                          & 5    & 24   \\
                                                   & Basic Theories of Psychological Development                         & 19   & 90   \\
                                                   & Biological Basis of Psychological Development and Fetal Development & 9    & 33   \\
                                                   & Intelligence                                                        & 9    & 34   \\
                                                   & Emotions                                                             & 8    & 34   \\
                                                   & Early Childhood Psychological Development                           & 36   & 114  \\
                                                   & Child Psychological Development                                 & 27   & 90   \\
                                                   & Adolescent Psychological Development                                & 18   & 60   \\
                                                   & Psychological Development in Adulthood                                     & 27   & 101  \\ \midrule
\multirow{4}{*}{Educational Psychology}            & Basic Concepts of Educational Psychology                            & 6    & 18   \\
                                                   & General Psychology of Learning                                         & 6    & 21   \\
                                                   & Major Theories of Learning                                          & 29   & 100  \\
                                                   & Categorical Learning Psychology                                     & 25   & 69   \\ \midrule \bottomrule
\end{tabular}
}
\caption{Details of the number of Concepts and Questions in each Chapter (Cont.)}
\label{tab:chapter-level-2}
\end{table*}
\section{Experiments Details of Calculating Concept Coverage Rate \label{sec: Experiments Details of Calculating Concept Coverage Rate}}
To calculate the concept coverage rate, the challenge lies in obtaining the required concepts in a subject. For advanced math, we first collect its chapters and the concepts under different chapters based on relevant exam requirements. We initially prompt GPT-4 to classify each question into different chapters, then prompt GPT-4 to further classify the question into specific instance-level concepts. We calculate the concept coverage rate and performance variance based on the concepts covered by these questions. This hierarchical classification allows for more accurate categorization.

For psychology in CMMLU, due to the vast number of concepts, prompting GPT-4 with these concepts in the input prompt each time is too expensive, so we use chapter-level concepts. We collect 84 chapter names according to the requirements, and classify each question into one or more chapters.

For other subjects, manually collecting concepts is too costly, and it's impossible to collect concepts for each subject. We prompt GPT-4 to generate a 3-level syllabus for the subject as the required concepts. We then prompt GPT-4 to categorize each question into one or more first-level headings in the syllabi. We further classify them into the second- and third-level headings under the selected first-level headings.
\section{More Results on Fine-grained Performance\label{sec: Concept level results for other models}}
\begin{figure*}[t!] 
\centering 
\includegraphics[width=\linewidth]{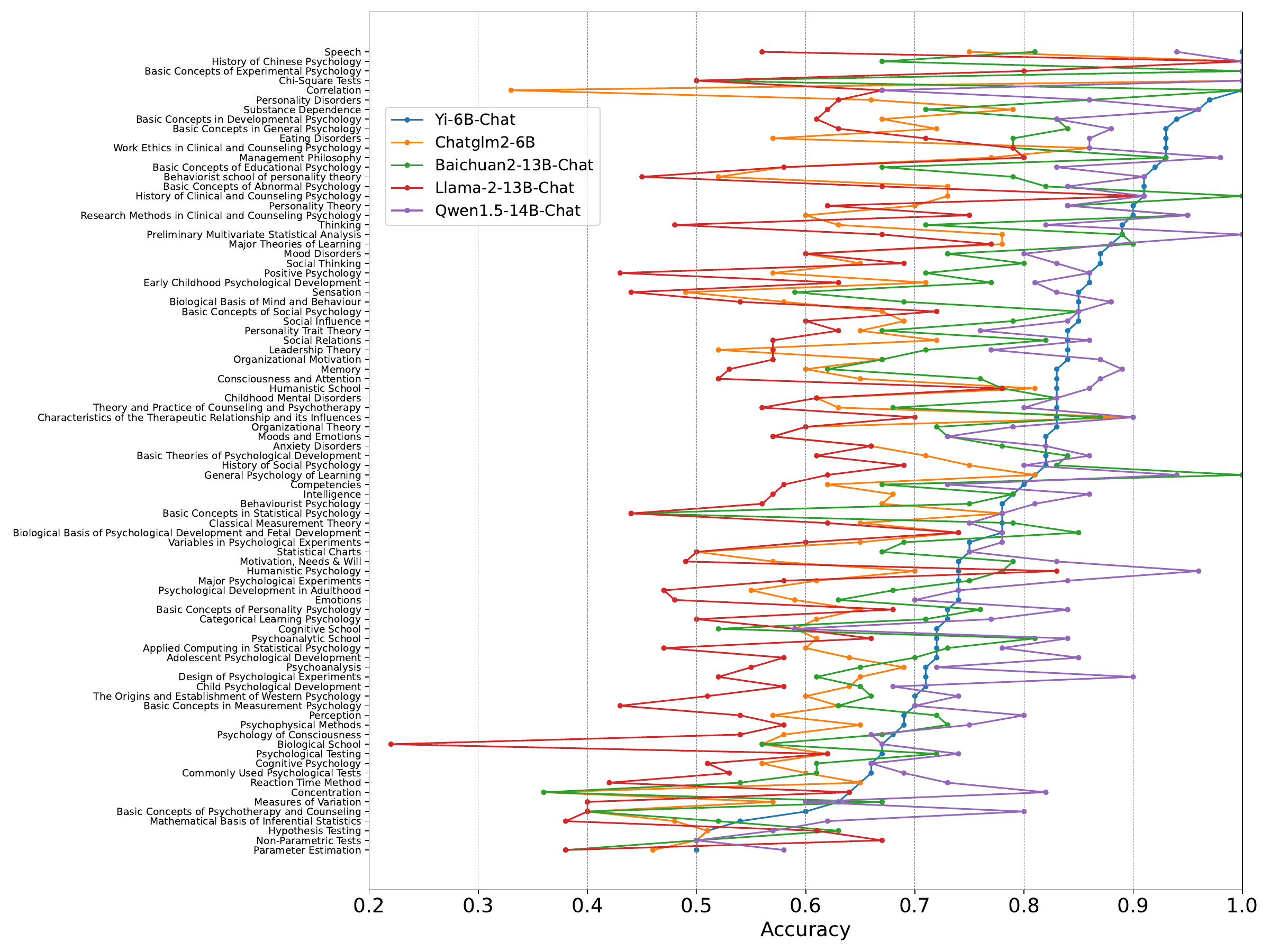} 
\caption{Concept-level results for models wi 14 Billion parameters }
\label{fig: conceptmap 14B} 
\end{figure*}

\begin{figure*}[t!] 
\centering 
\includegraphics[width=\linewidth]{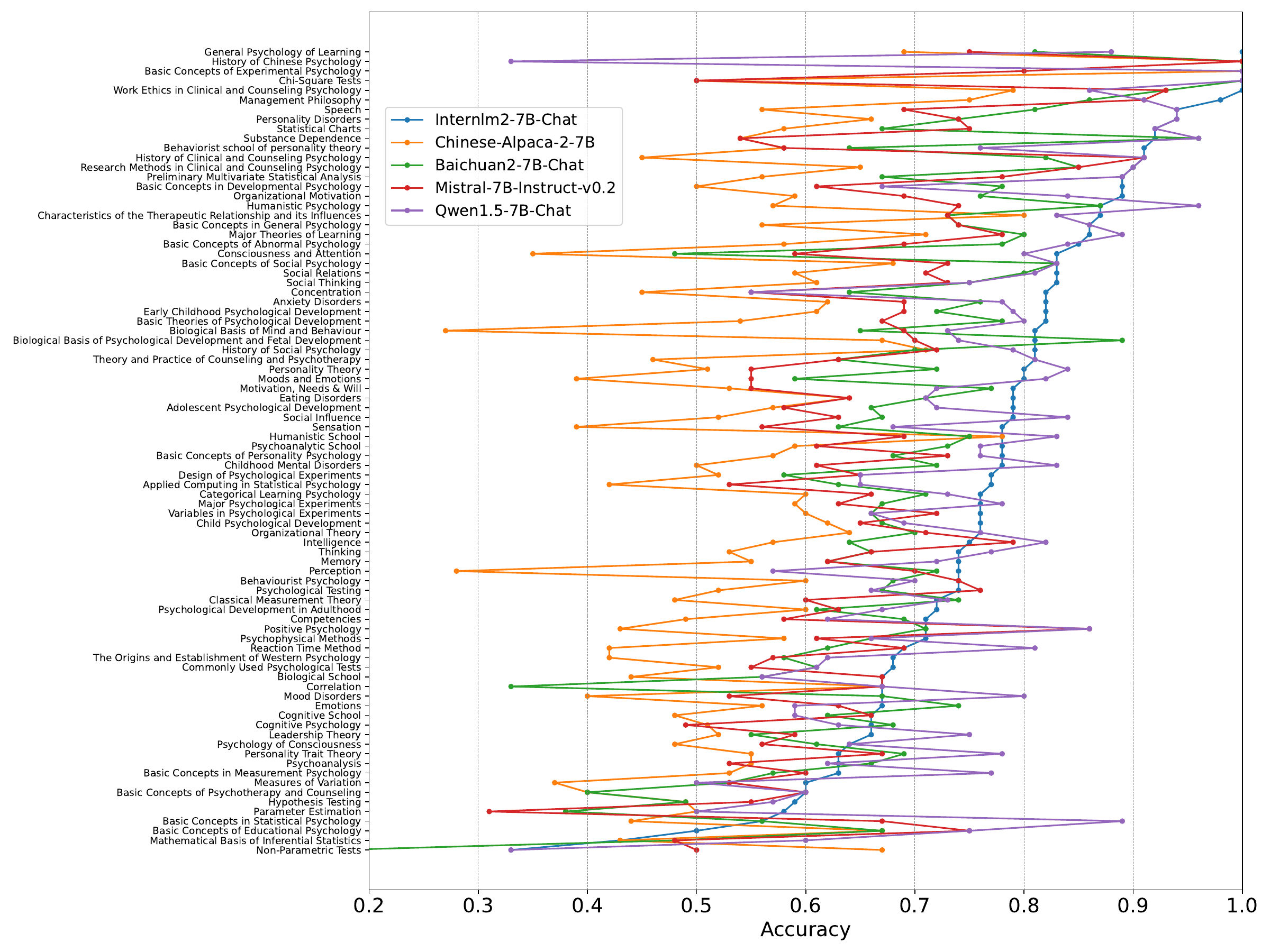} 
\caption{Concept-level results for some other models}
\label{fig: conceptmap other} 
\end{figure*}

We provide more chapter-level results in this Figure \ref{fig: conceptmap 14B}, \ref{fig: conceptmap other}.
\section{Qualifications  of Our psychology Annotators}
Our psychological annotators is professional. We enlist three psychological annotators with diverse expertise. The first is a postdoctoral researcher specializing in experimental and counseling psychology. The second is a registered psychology researcher with the British Psychological Society with five years of research experience in interdisciplinary psychology laboratories. The third, a counseling psychologist, brings over three years of practical counseling experience.

\section{Details About Evaluated Models \label{sec: Details About Evaluated Models}}

\begin{table*}[t!]
\centering
\begin{tabular}{ll}
\toprule
\textbf{Model Name} & \textbf{Model Code/API} \\
\midrule
\texttt{Chinese-Alpaca-2-7B}\cite{Chinese-LLaMA-Alpaca} & hfl/chinese-alpaca-2-7b \\ 
\texttt{Chatglm2-6B}\cite{du2022glm} & THUDM/chatglm2-6b \\
\texttt{Llama-2-13B-Chat}\cite{touvron2023llama} & meta-llama/Llama-2-13b-chat-hf \\
\texttt{Llama-2-70B-Chat}\cite{touvron2023llama} 
& meta-llama/Llama-2-70b-chat-hf \\
\texttt{Baichuan2-7B-Chat}\cite{yang2023baichuan} & baichuan-inc/Baichuan2-7B-Chat \\
\texttt{Baichuan2-13B-Chat}\cite{yang2023baichuan} & baichuan-inc/Baichuan2-13B-Chat \\
\texttt{Mistral-7B-Instruct-v0.2}\cite{jiang2023mistral} & mistralai/Mistral-7B-Instruct-v0.2 \\
\texttt{Mixtral-8x7B-Instruct-v0.1}\cite{jiang2024mixtral} & mistralai/Mixtral-8x7B-Instruct-v0.1 \\
\texttt{Qwen1.5-7B-Chat}\cite{bai2023qwen} & Qwen/Qwen1.5-7B-Chat \\
\texttt{Qwen1.5-72B-Chat}\cite{bai2023qwen} & Qwen/Qwen1.5-72B-Chat \\
\texttt{Qwen1.5-14B-Chat}\cite{bai2023qwen} &Qwen/Qwen1.5-14B-Chat \\
\texttt{Internlm2-7B-Chat}\cite{cai2024internlm2} & internlm/internlm2-chat-7b \\
\texttt{Yi-6B-Chat}\cite{ai2024yi} & 01-ai/Yi-6B-Chat \\
\texttt{Yi-34B-Chat}\cite{ai2024yi} & 01-ai/Yi-34B-Chat \\
\texttt{GPT-3.5-Turbo}\cite{gpt-3.5} & Azure api: gpt-35-turbo \\
\bottomrule
\end{tabular}
\caption{ Model code/API of our evaluated models.}
\label{tab:model list}
\end{table*}

The evaluated models include \texttt{Chinese-Alpaca-2-7B}\cite{Chinese-LLaMA-Alpaca}, \texttt{Chatglm-6B}\cite{du2022glm}, \texttt{Chatglm2-6B}\cite{du2022glm}, \texttt{Llama-2-13B-Chat}\cite{touvron2023llama}, \texttt{Llama-2-70B-Chat}\cite{touvron2023llama}, \texttt{Baichuan2-7B-Chat}\cite{yang2023baichuan}, \texttt{Baichuan2-13B-Chat}\cite{yang2023baichuan}, \texttt{Mistral-7B-Instruct-v0.2}\cite{jiang2023mistral}, \texttt{Mixtral-8x7B-Instruct-v0.1}\cite{jiang2024mixtral}, \texttt{Qwen1.5-7B-Chat}\cite{bai2023qwen}, \texttt{Qwen1.5-72B-Chat}\cite{bai2023qwen}, \texttt{Qwen1.5-14B-Chat}\cite{bai2023qwen}, \texttt{Internlm2-7B-Chat}\cite{cai2024internlm2}, \texttt{Yi-6B-Chat}\cite{ai2024yi}, and \texttt{Yi-34B-Chat}\cite{ai2024yi}. The model codes can be found in Table \ref{tab:model list} . We evaluate these models with vLLM~\citep{kwon2023efficient}.

\section{Prompts for Questions Generation \label{appendix:Prompts for Questions Generation}}
We totally design four distinct prompts to steer GPT-4 in generating questions based on provided knowledge points. In designing these prompts, we have taken into account several guidelines. Firstly, the generated questions should exhibit a high level of difficulty and complexity. Secondly, while generating questions, GPT-4 should primarily rely on the given knowledge points, but it can also incorporate its inherent psychological knowledge. Thirdly, each knowledge point unit may yield multiple questions, but the content, type, or perspective of the questions should be distinct.

\begin{figure*}[t!]
\centering
  \includegraphics[width=1.0\linewidth]{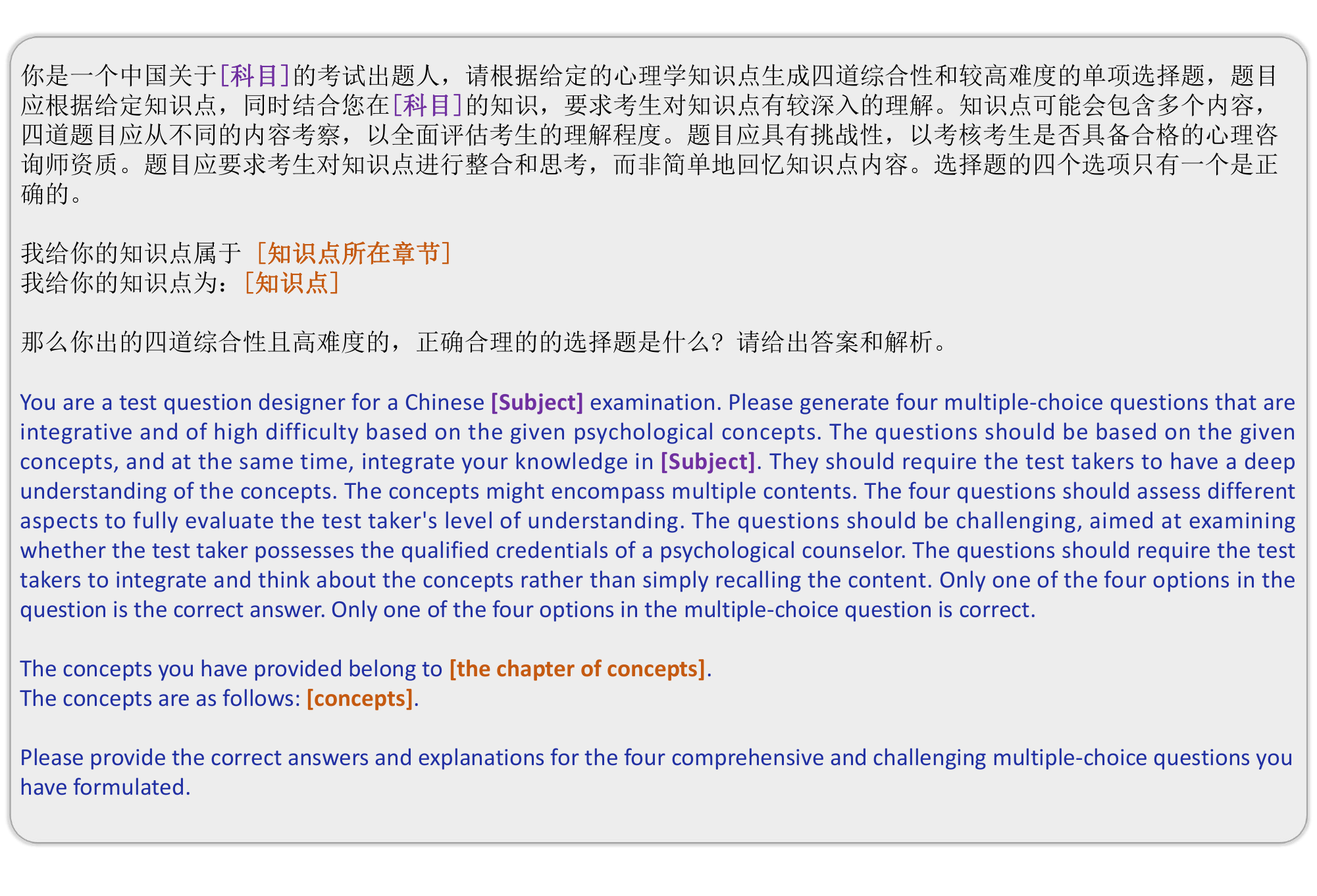}
  \caption{Theory understanding }
  \label{fig: Theory_understanding_prompt}
\end{figure*}

\begin{figure*}[t!]
\centering
  \includegraphics[width=1.05\linewidth]{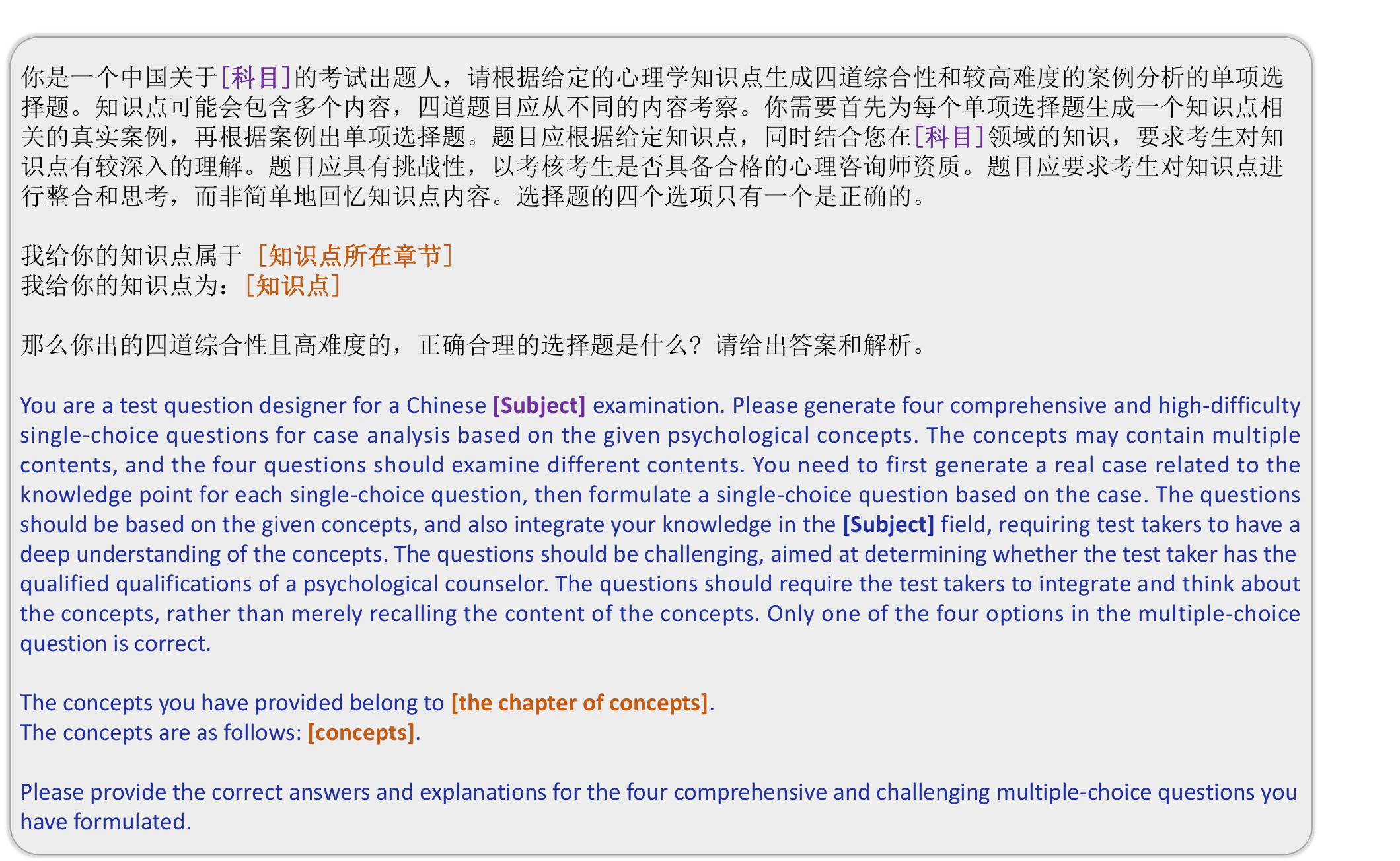}
  \caption{Case study }
  \label{fig: case_study_prompt}
\end{figure*}

\begin{figure*}[t!]
\centering
  \includegraphics[width=1.0\linewidth]{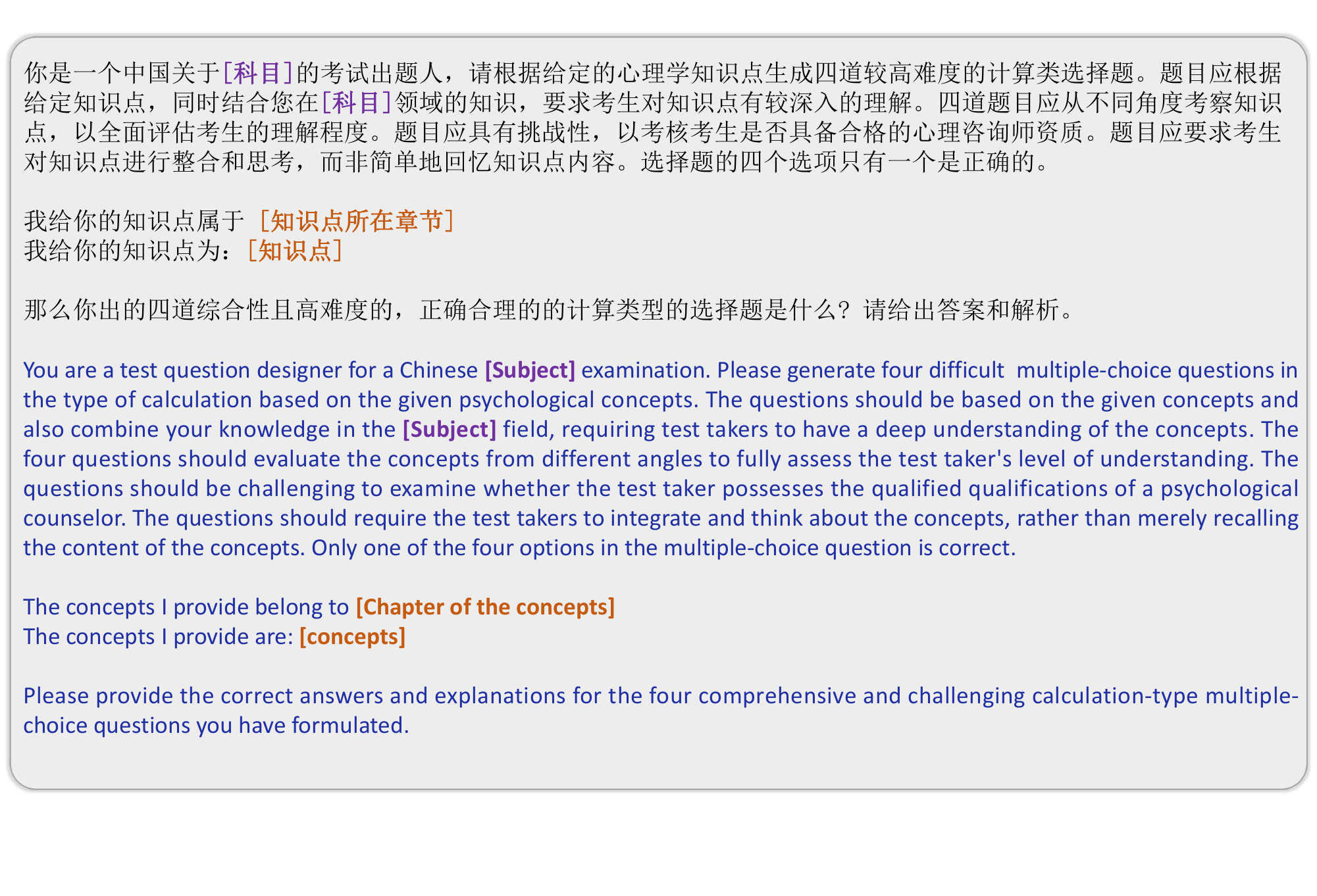}
  \caption{Calculation }
  \label{fig: cal_prompt_full}
\end{figure*}

\begin{figure*}[t!]
\centering
  \includegraphics[width=1.0\linewidth]{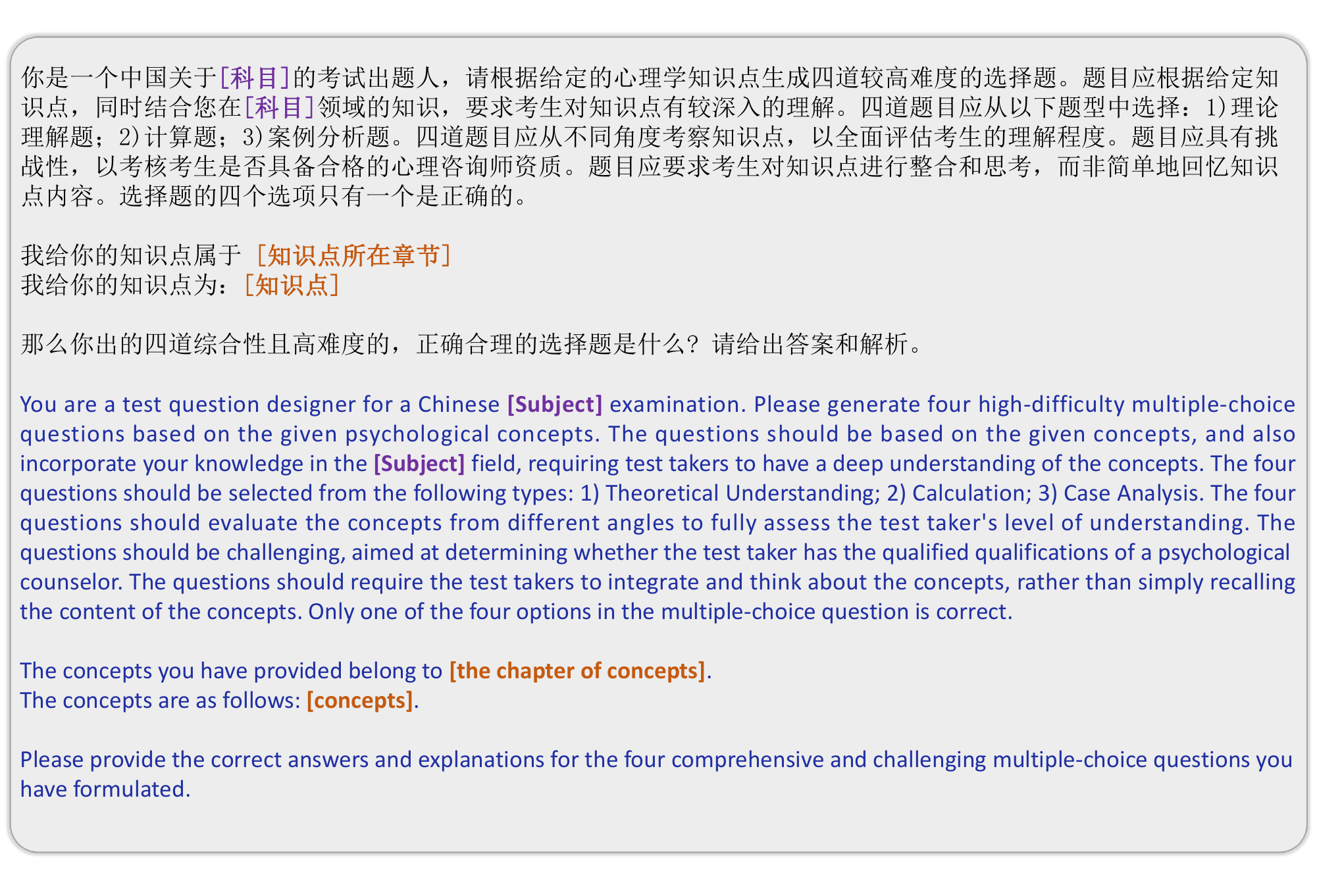}
  \caption{Multiple types prompt}
  \label{fig: multiple_type_prompt}
\end{figure*}

Figures \ref{fig: Theory_understanding_prompt}, \ref{fig: case_study_prompt}, \ref{fig: cal_prompt_full}, and \ref{fig: multiple_type_prompt} show the specific prompts that we have inputted into GPT-4. These prompts have been meticulously designed, with each one being tailored to control the generation of a different kind of question. The first three prompts correspond to generating Theory understanding, Case study, and Calculation type questions, respectively, while the last one encompasses all of the aforementioned types. We choose the appropriate prompt for each knowledge point based on its content.

\begin{figure*}[t!]
\centering
  \includegraphics[width=\linewidth]{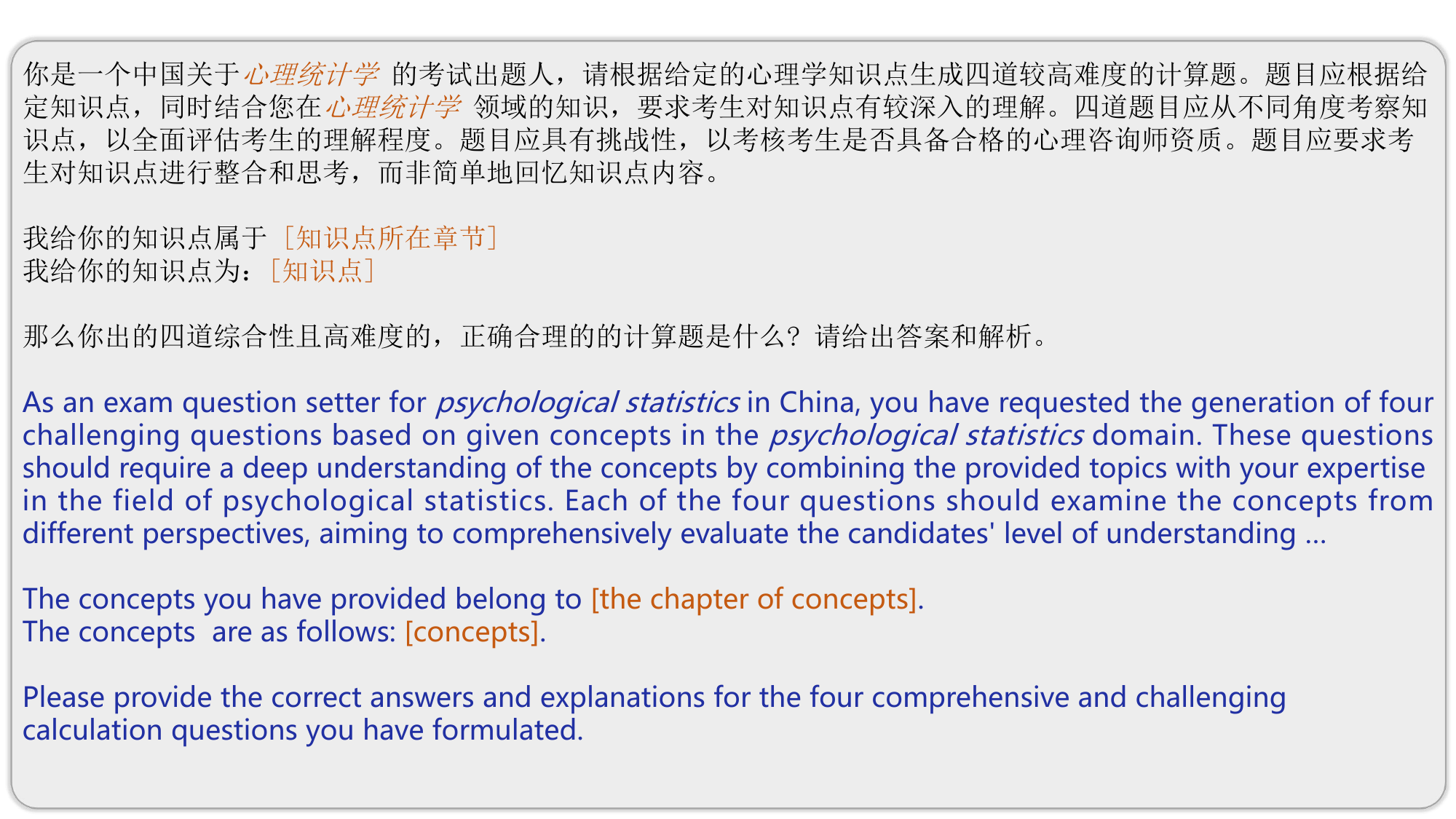}
  \caption{The question generation prompt template (translated in English), which is primarily designed for generating the type of calculation questions. }
  \label{fig:cal_prompt}
\end{figure*}
\begin{figure*}[t!]
\centering
  \includegraphics[width=\linewidth]{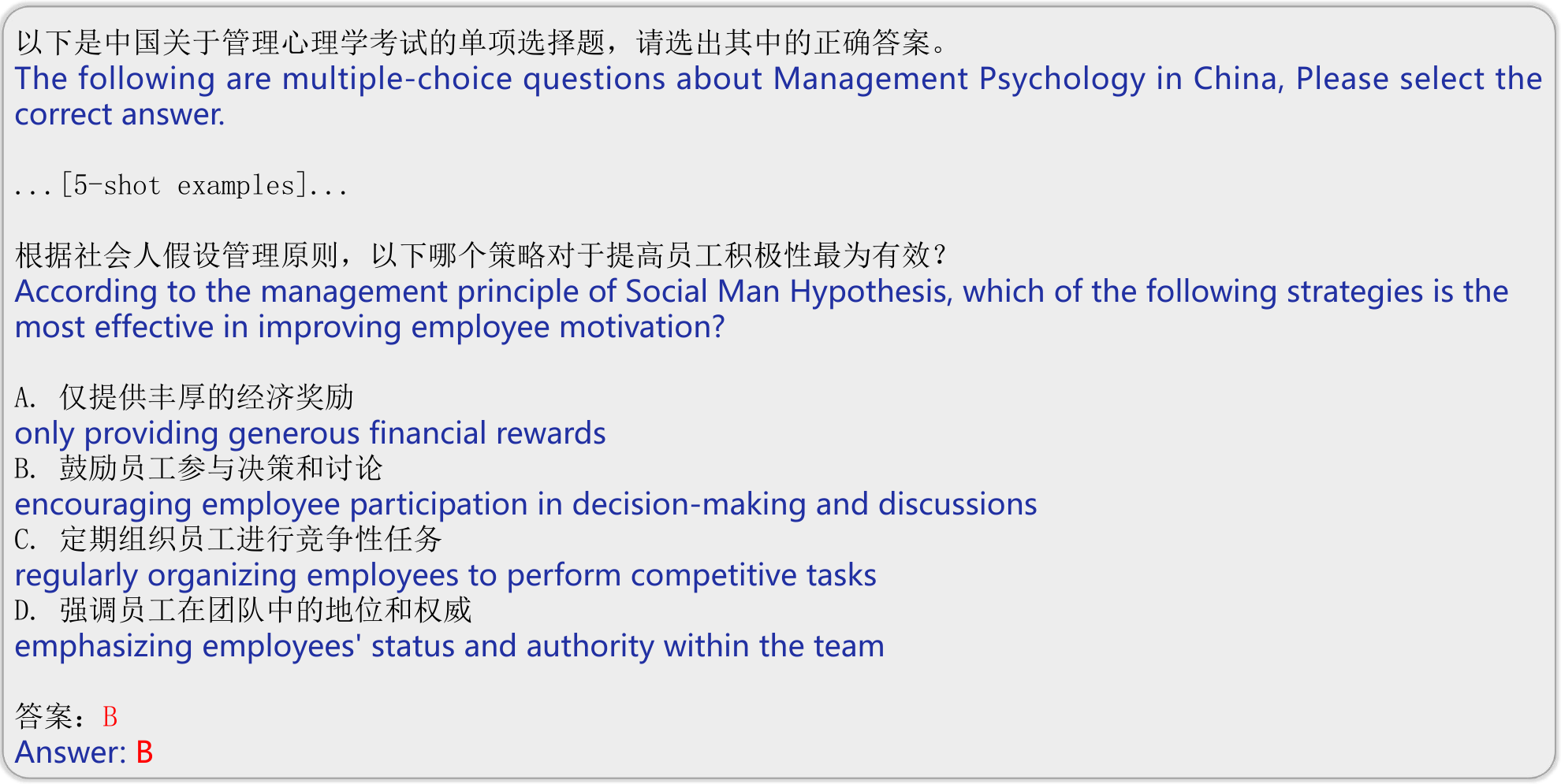}
  \caption{An example of prompts in few-shot setting. The black text is what we feed into model, while the red text is the response completed by model. The English translation for the Chinese input is provided in the purple text, which is not included in the actual prompt. }
  \label{fig:prompteval}
\end{figure*}

\end{document}